\documentclass[a4paper]{amsart}

\usepackage{amsmath,amsfonts}
\usepackage{algorithmic}
\usepackage{algorithm}
\usepackage{array}
\usepackage{textcomp}
\usepackage{stfloats}
\usepackage{url}
\usepackage{verbatim}
\usepackage{graphicx}
\usepackage{cite}

\usepackage{geometry}
\usepackage{amsmath, amssymb, amsfonts, amsthm, amscd, amsopn, amstext, amsxtra, euscript}

\usepackage{graphicx}
\usepackage{xcolor}
\usepackage{tikz}
\usetikzlibrary{arrows.meta, positioning}
\usepackage{xy}
\xyoption{all}

\usepackage[utf8]{inputenc}

\usepackage{longtable}
\usepackage{enumitem}
\usepackage{booktabs}
\usepackage{algorithm}
\usepackage{algorithmic}

\usepackage{hyperref}
\hypersetup{
  colorlinks   = true,
  urlcolor     = blue,
  linkcolor    = blue,
  citecolor    = blue
}
\usepackage[capitalise]{cleveref}

\usepackage[citation-order, msc-links]{amsrefs}

\newtheorem{thm}{Theorem}[section]
\newtheorem{prop}[thm]{Proposition}
\newtheorem{lem}[thm]{Lemma}
\newtheorem{cor}[thm]{Corollary}

\newtheorem{df}[thm]{Definition}

\newtheorem{prop-alg}[thm]{Proposed Algorithm}
\newtheorem{ex}[thm]{Example}
\newtheorem{rem}[thm]{Remark}
\newtheorem{algocf}{Algorithm}[section]

\crefname{thm}{Thm.}{}
\crefname{prop}{Prop.}{}
\crefname{lem}{Lem.}{}
\crefname{cor}{Cor.}{}
\crefname{algocf}{Algorithm}{}
\crefname{item}{Item}{}
\crefname{df}{Def.}{}
\crefname{section}{Sec.}{}

\newcommand{\F}{\mathbb{F}}

\newcommand{\R}{\mathbb{R}}
\newcommand{\N}{\mathbb{N}}
\newcommand{\Z}{\mathbb{Z}}
\newcommand{\Q}{\mathbb{Q}}

\newcommand{\cA}{\mathcal{A}}
\newcommand{\cB}{\mathcal{B}}
\newcommand{\cF}{\mathcal{F}}
\newcommand{\cL}{\mathcal{L}}
\newcommand{\cS}{\mathcal{S}}
\newcommand{\cT}{\mathcal{T}}

\newcommand{\V}{\mathcal{V}}     				

\newcommand{\Voc}{\mathsf{Voc}}

\newcommand{\w}{\mathbf{q}}
\newcommand{\x}{\mathbf{x}}
\newcommand{\y}{\mathbf{y}}
\newcommand{\z}{\mathbf{z}}
\newcommand{\h}{\mathbf{h}}
\newcommand{\bk}{\mathbf{k}}
\newcommand{\bv}{\mathbf{v}}
\newcommand{\bb}{\mathbf{b}}
\newcommand{\bsc}{\mathbf{sc}}
\newcommand{\bsf}{\mathbf{smax}}     
\newcommand{\W}{\mathbf{W}}
\newcommand{\pos}{\mathbf{t}}

\DeclareMathOperator{\Lgt}{\Psi_\w}

\DeclareMathOperator{\EGT}{{\bar \Psi}_{\w,\lambda}}

\newcommand{\Mat}{\operatorname{Mat}}
\newcommand{\fnn}{\operatorname{Fnn}}
\newcommand{\Ln}{\operatorname{Ln}}
\newcommand{\En}{\operatorname{En}}
\newcommand{\Dcl}{\operatorname{Dcl}}
\newcommand{\Dc}{\operatorname{Dc}}
\newcommand{\Mh}{\operatorname{Mh}}
\newcommand{\Ca}{\operatorname{Ca}}
\newcommand{\Mmh}{\operatorname{Mmh}}

\newcommand{\EnL}{\operatorname{EnL}}
\newcommand{\Ct}{\operatorname{Ct}}
\newcommand{\Head}{\operatorname{Head}}
\newcommand{\relu}{\operatorname{ReLU}}
\newcommand{\vc}{\operatorname{vc}}
\newcommand{\diag}{\operatorname{diag}}
\newcommand{\pe}{\operatorname{Pe}}
\newcommand{\pep}{\operatorname{Pe}^\prime}

\DeclareMathOperator{\deff}{d_{\mathrm{eff}}}

\DeclareMathOperator{\Mlw}{M_{\w}}
\DeclareMathOperator{\Llw}{L_{\w}}
\DeclareMathOperator{\Mw}{M_{\w,\lambda}}
\DeclareMathOperator{\Megt}{M_{\w_i,\la}}

\DeclareMathOperator{\Lw}{L_{\w,\lambda}}
\DeclareMathOperator{\gr}{gr}

\DeclareMathOperator{\tr}{tr}

\newcommand{\la}{\lambda}

\newcommand\Lip{\Lambda}   

\newcommand{\norm}[1]{\Vert#1\Vert}

\def\Lmse{\mathfrak{L}_{\text{MSE}}} 		
\def\Lnorm{\mathfrak{L}_{\text{norm}}} 	
\def\Lhom{\mathfrak{L}_{\text{hom}} } 	
\def\Lce{\mathfrak{L}_{\text{ce}}}  		
\def\Lmax{\mathfrak{L}_{\text{max}}} 		

\usepackage{tikz}
\usetikzlibrary{arrows.meta,positioning,fit}

\tikzset{
  gt/block/.style={
    draw,
    rounded corners=2pt,
    minimum width=0.9\columnwidth,
    minimum height=1.2cm,
    inner sep=6pt,
    fill=black!10
  },
  gt/arrow/.style={
    -{Latex[length=2.5mm]},
    line width=0.6pt
  },
  gt/side/.style={
    align=left,
    text width=0.9\columnwidth
  }
}

\title[Graded Transformers]{Graded Transformers}

\author{T. Shaska}
\address{Department of Mathematics and Statistics, \\ Oakland University, Rochester, MI 48309}
\email{shaska@oakland.edu}

\dedicatory{To my parents, who never bowed, and never stopped teaching.}


\begin{document}

\begin{abstract} 
We introduce the Graded Transformer framework, embedding algebraic inductive biases via grading transformations on vector spaces. Extending Graded Neural Networks (GNNs), we propose the Linearly Graded Transformer (LGT) and Exponentially Graded Transformer (EGT), which apply parameterized scaling—via fixed or learnable grading tuples, with exponential factors for EGT—to encode hierarchies in attention and representation layers. We establish rigorous guarantees, including universal approximation, reduced sample complexity via VC dimension bounds, Lipschitz continuity, and robustness to perturbations. A graded loss ensures stable optimization, with learnable grades enabling adaptive feature scaling. The framework excels in hierarchical and neuro-symbolic tasks across algebraic geometry, physics, NLP, biological sequence analysis, and emerging domains like graphs and finance, offering efficiency and interpretability over data-driven models. By integrating geometric and algebraic principles, Graded Transformers provide a principled approach to structured deep learning, with potential to reshape modeling of complex systems.  
\end{abstract}
 
\maketitle
 

\section{Introduction}
\label{sec-1}

Our prior work on AI over graded vector spaces in \cite{2024-2} and Graded Neural Networks (GNNs) in \cite{sh-89} established a framework for embedding algebraic structure into neural architectures, using fixed numerical grades to prioritize hierarchical features, enhancing efficiency and interpretability in domains like algebraic geometry and photonic signal processing. This paper introduces the \emph{Graded Transformer} framework, comprising the Linearly Graded Transformer (LGT) and Exponentially Graded Transformer (EGT), which build on \cite{2024-2,sh-89} by integrating fixed-grade principles with the dynamic, context-aware learning of transformers. The Graded Transformer employs grading transformations \(\Llw\) (LGT) or \(\Lw\) (EGT)   to prioritize critical features, with LGT using linear weights and EGT using exponential scaling. An extension to learnable grades   enhances adaptability for complex, hierarchical data across scientific and linguistic domains.

Sequence modeling underpins modern machine learning, enabling breakthroughs in natural language processing (NLP), time-series analysis, and biological sequence analysis by capturing long-range dependencies. The transformer architecture has revolutionized this field through self-attention, achieving state-of-the-art performance in tasks like machine translation, physical simulations, and genomic analysis. However, transformers struggle with hierarchical data structures, such as polynomial degrees in algebraic geometry, multi-scale phenomena in physics, syntactic heads in NLP, or regulatory regions in biology, requiring extensive training data to uncover domain-specific patterns; see \cite{yun2019transformers}. This leads to high sample complexity, increased computational costs, and limited interpretability when hierarchical relationships are known a priori. The Graded Transformer addresses these challenges by embedding algebraic biases into attention and representation layers, prioritizing critical features without relying solely on data-driven attention.

The growing prevalence of hierarchical data, from algebraic systems to large language models (LLMs), underscores the need for models that leverage structural priors to enhance efficiency and interpretability. Unlike structured attention mechanisms or graph neural networks, which often sacrifice flexibility or require complex preprocessing (see \cite{vaswani2017attention} for details), the Graded Transformer, rooted in the GNN framework of \cite{sh-89}, offers a compelling alternative. By encoding domain knowledge through grading tuples, it pursues three primary objectives:

i)  \emph{Feature Prioritization}: Highlighting significant features to reduce dependence on large datasets, enabling efficient learning in structured domains.
  
ii)  \emph{Computational Efficiency}: Leveraging structural priors to lower sample complexity, enabling faster convergence for hierarchical tasks.
  
iii) \emph{Interpretability}: Encoding domain knowledge transparently via grading tuples, making the model’s behavior predictable and explainable.

This paper is organized as follows:
In     \cref{sec-2}   we develop the algebraic foundation of the framework by formalizing graded vector spaces, where basis vectors are assigned numerical grades that reflect structural importance. It introduces scalar actions, graded linear maps, and their invariance properties, laying the groundwork for encoding inductive biases into neural networks. The notion of grading is generalized to allow fractional, integer, or exponential scales, and used to define norms and activations compatible with the algebraic structure of the space.

\cref{sec-3}   defines additive and multiplicative graded neurons and introduces graded activation functions and loss functions that respect the underlying grading geometry. It formalizes graded layers and networks, highlighting how scaling via grades allows prioritization of critical features. The section also introduces graded versions of ReLU and cross-entropy loss, showing how standard architectures can be adapted to operate on graded spaces with stable and interpretable behavior.

\cref{sec-4} presents the classical transformer architecture as a baseline, detailing its components — embedding, attention, encoder, decoder, and autoregressive generation — with an emphasis on permutation equivariance and computational complexity. It analyzes the transformer's limitations in modeling hierarchical data, referencing its high sample complexity and uniform attention behavior. These deficiencies motivate the graded extensions introduced in later sections, particularly in domains where structure is known in advance.

In \cref{sec-5} the Linearly Graded Transformer is defined by applying a fixed diagonal scaling transformation to the input representation before applying the standard transformer. The section proves universal approximation, derives VC dimension bounds, and shows robustness to noise. Graded attention mechanisms are introduced, and it is shown that linear grading preserves algebraic invariance, enhances attention rank, and improves training efficiency. The architecture preserves the computational complexity of standard transformers while improving structure-awareness and interpretability.

\cref{sec-6}  extends the framework to exponentially graded transformations, enabling stronger amplification of high-grade features. It formalizes the exponential grading map and proves an analogous universal approximation theorem and robustness results. The section highlights use cases where exponential growth captures deep hierarchies, such as symbolic expressions or physical systems with scale separation. Attention expressivity is shown to be enhanced via exponential weighting, providing finer control over attention focus.

In \cref{sec-7} the training dynamics of graded transformers are addressed, focusing on stability, gradient flow, and optimization strategies for both fixed and learnable grades. Graded loss functions are revisited in the context of back propagation, and new training algorithms are introduced to accommodate dynamically updated grading tuples. The section also discusses regularization schemes, gradient normalization, and convergence behavior in the graded setting, providing a principled route toward training stable and efficient models.

The final \cref{sec-8} demonstrates the versatility of graded transformers through applications in diverse domains. In algebraic geometry, they model polynomial structures and zeta functions; in physics, they capture multi-scale and turbulent phenomena; in NLP, they highlight syntactic heads and semantic roles; in biology, they prioritize functional subsequences in proteins and DNA. Emerging domains like graph learning and symbolic reasoning are also discussed, underscoring the framework’s capacity to bridge data-driven learning with mathematical structure.

The Graded Transformer stands poised to revolutionize artificial intelligence and scientific discovery by blending algebraic structure with data-driven flexibility, building on our foundational work in \cite{2024-2,sh-89,sh-97}. For LLMs, it offers a transformative approach to prioritize critical tokens or syntactic structures, reducing computational demands while enhancing performance in tasks like dialogue understanding, question answering, automated reasoning, and multi-modal learning. In scientific domains, its algebraic grounding, rooted in fixed and learnable grades, enables data-efficient solutions for challenges like computing zeta functions in algebraic geometry, modeling turbulent flows in physics, or predicting protein structures in biology, as detailed in \cref{sec-8}. This fusion of mathematical rigor and adaptability invites readers to explore a new paradigm of structured learning, promising interpretable, efficient, and mathematically principled models for the complex realities of the world.

\section{Graded Vector Spaces}\label{sec-2}
Here we provide the essential background on graded vector spaces, extended to incorporate recent advancements in their structure and operations. The interested reader can check details at     \cite{2024-2} which will be our main reference for this part. We use "grades" to denote the indices of grading (e.g., $q_i$), distinguishing them from "weights" used for neural network coefficients.

A graded vector space equips subspaces with numerical grades, enabling differential scaling of components to reflect their relative importance. 
This algebraic structure underpins the graded transformer by embedding numerical grades into attention mechanisms and loss functions, prioritizing features like high-degree terms in algebraic geometry or critical tokens in natural language processing.

\index{graded vector space}
\index{grading tuple}
 \index{grading transformation}

\begin{df} 
Let \( \F \) be a field, and let \( V \) be a vector space over \( \F \) with basis \( \cB = \{e_0, \ldots, e_{d-1}\} \). A \textbf{graded vector space}  is a direct sum decomposition
\[
V = \bigoplus_{i=0}^{d-1} V_i,
\]
where \( V_i = \F e_i \) are the one-dimensional subspaces spanned by \( e_i \), assigned a grade \( q_i \in \R \). 
\end{df}

In this paper, we often assume $q_i \in \Q_{>0}$ for positive scaling in machine learning tasks (see \cite{2024-2}).
The grades form the \textbf{grading tuple} \( \w = (q_0, \ldots, q_{d-1}) \). A vector \( \x \in V \) is expressed as
\[
\x = \sum_{i=0}^{d-1} x_i e_i, \quad x_i \in \F,
\]
with the component \( x_i e_i \in V_i \) having grade \( q_i \).

We will denote by $\V_{\w}(\F)$  a graded vector space over $\F$ with grading tuple $\w$. 
There is a large amount of literature on graded modules and graded vector spaces. The interested reader can follow the references from \cite{2024-2}.  Below we present a bare minimum of terminology that is needed for this paper. 

\subsection{Generalized Gradation}
Let $I$ be an index set, which may be $\N$, $\Z$, a field like $\Q$, or a monoid. An \textbf{$I$-graded vector space}   $V$ is a vector space with a decomposition:
$$V = \bigoplus_{i \in I} V_i,$$
where each $V_i$ is a vector space. An element $v \in V$ is called \textbf{homogeneous of degree $i$} if it belongs to $V_i$ (i.e., it has no components in other subspaces $V_j$ for $j \neq i$).
When $I = \Q$, grades can represent fractional weights, useful for modeling continuous hierarchies in machine learning tasks as in \cite{2024-2}. For $I = \N$, we recover the standard $\N$-graded vector space, often simply called a \textbf{graded vector space}.

\begin{ex}
For a $\Q$-graded space with grades $q_0 = 1, q_1 = 1/2$, a vector $[x_0, x_1]$ assigns higher priority to $x_0$, useful for scaling features in neural networks.
\end{ex}
 
\subsection{Graded Linear Maps}
For an index set $I$, a linear map $f: V \to W$ between $I$-graded vector spaces is a \textbf{graded linear map} if it preserves the grading, i.e., $f(V_i) \subseteq W_i$ for all $i \in I$. Such maps are also called \textbf{homomorphisms} (or \textbf{morphisms}) of graded vector spaces or \textbf{homogeneous linear maps of degree 0}. 

For a commutative monoid $I$ (e.g., $\N$), a map $f$ is \textbf{homogeneous of degree $d \in I$} if
\[
f(V_j) \subseteq W_{j+d}, \quad \text{for all } j \in I,
\]
where $+$ is the monoid operation. If $I$ is an additive group (e.g., $\Z$ or $\Q$), maps of degree $d \in I$ shift grades similarly, including negative or fractional degrees. For example, a map of degree $d$, where $d = -k$ for some $k \in I$, satisfies
\[
f(V_{j+k}) \subseteq W_j, \quad f(V_j) = 0 \text{ if } j - k \notin I,
\]
reducing the grade by $k$. This is particularly relevant for $\Q$-graded spaces in machine learning, where negative or fractional grades enable flexible feature prioritization, as detailed in \cite{2024-2}.

Let $\V^n_\w(\F)$ and $\V^m_{\w'}(\F)$ be graded vector spaces with $\w = (q_0, \ldots, q_{n-1})$ and $\w' = (r_0, \ldots, r_{m-1})$, respectively, where $q_i, r_j \in \Q_{>0}$. 
Let 
\[
\phi : \V^n_\w(\F) \to \V^m_{\w'}(\F)
\]
be an $\F$-linear map and $A = (a_{ij}) \in \Mat_{m \times n}(\F)$ its associated matrix with respect to the standard bases. A map $\phi$ is \textbf{grade-preserving} if it is homogeneous of degree 0, i.e., $\phi(V_i) \subseteq W_i$ for all $i \in I$. Then we have the following: 

\begin{lem}
$\phi$ is homogeneous of degree $d \in \Q$ if and only if
\[
a_{ij} \neq 0 \quad \Longrightarrow \quad r_i = q_j + d.
\]
In particular, $\phi$ is grade-preserving if and only if $a_{ij} \neq 0$ implies $r_i = q_j$.
\end{lem}

\begin{proof}
Assume \(\phi\) is homogeneous of degree \(d \in \Q\). 
For a basis vector \(e_j \in V_j\) of grade \(q_j\), \(\phi(e_j)\) must lie in the subspace of grade \(q_j + d\), i.e., \(\phi(e_j) \in \bigoplus_{i: r_i = q_j + d} W_i\). Since \(\phi(e_j) = \sum_{i=0}^{m-1} a_{ij} f_i\), where \(A = (a_{ij})\) is the matrix of \(\phi\), non-zero components \(a_{ij} \neq 0\) occur only when \(f_i \in W_i\) has grade \(r_i = q_j + d\). Thus, $a_{ij} \neq 0$  implies $r_i = q_j + d$.

Conversely, 
for a basis vector \(e_j \in V_j\) of grade \(q_j\), we have
\[
\phi(e_j) = \sum_{i=0}^{m-1} a_{ij} f_i = \sum_{i: r_i = q_j + d} a_{ij} f_i,
\]
since \(a_{ij} = 0\) when \(r_i \neq q_j + d\). Thus, \(\phi(e_j)\) lies in \(\bigoplus_{i: r_i = q_j + d} W_i\), which has grade \(q_j + d\). 
For any  \(\x = \sum_{j=0}^{n-1} x_j e_j \in \V^n_\w(\F)\), where \(x_j e_j \in V_j\) has grade \(q_j\), we have 
\[
\phi(\x) = \sum_{j=0}^{n-1} x_j \phi(e_j) = \sum_{j=0}^{n-1} x_j \sum_{i: r_i = q_j + d} a_{ij} f_i.
\]
Each term \(x_j \phi(e_j)\)  is in \(\bigoplus_{i: r_i = q_j + d} W_i\), so \(\phi(\x)\) is a sum of homogeneous components of grade \(q_j + d\) for each \(q_j\) present in \(\x\). Hence, \(\phi\) is homogeneous of degree \(d\).

The  case, \(\phi\) is grade-preserving (degree 0) if and only if \(\phi(V_j) \subseteq W_j\), i.e., \(d = 0\). Substituting \(d = 0\) into the condition, we get \(a_{ij} \neq 0 \implies r_i = q_j\), as required.
\end{proof}

Graded scalar multiplication $\star : \F \times \V^n_\w(\F) \to \V^n_\w(\F)$, given by 
\[
\lambda \star \x = (\lambda^{q_0} x_0, \ldots, \lambda^{q_{n-1}} x_{n-1})
\]
 for $\lambda \in \F$, is a graded linear map of degree 0.

\begin{lem}[Grading-Invariant Subspaces]\label{grad-inv}
Let \(W \subseteq \V^n_\w(\F)\) be an \(\F\)-linear subspace. The following are equivalent:
\begin{enumerate}[label=\roman*), leftmargin=*]
    \item \(W\) is invariant under scalar action: \(\lambda \star \x \in W\) for all \(\lambda \in \F^\times\) and all \(\x \in W\).
    \item \(W\) is generated by homogeneous vectors in \(\V^n_\w(\F)\).
\end{enumerate}
\end{lem}

\begin{proof}
Assume \(W\) is invariant under \(\lambda \star \x\). For \(\x = \sum_{j=0}^{n-1} x_j e_j \in W\), we have \(\lambda \star \x = \sum_{j=0}^{n-1} \lambda^{q_j} x_j e_j \in W\). Since grades \(q_j \in \Q_{>0}\) are distinct, choose \(\lambda_1, \ldots, \lambda_n \in \F^\times\) such that \([\lambda_i^{q_j}]\) is invertible. Solving the system \(\sum_{j=0}^{n-1} \lambda_i^{q_j} x_j e_j \in W\) yields \(x_j e_j \in W\), so \(W\) is spanned by homogeneous vectors.

Conversely, if \(W = \text{span}\{x_j e_j \mid j \in J\}\), each \(x_j e_j \in V_j\). For \(\x = \sum_{j \in J} c_j (x_j e_j) \in W\), we have \(\lambda \star \x = \sum_{j \in J} c_j \lambda^{q_j} (x_j e_j) \in W\), since each \(\lambda^{q_j} (x_j e_j) \in V_j\). Thus, \(W\) is invariant.
\end{proof}

\begin{cor}\label{cor:grading_invariant}
Each \(V_j = \F e_j\) of grade \(q_j \in \Q_{>0}\) is a maximal subspace of \(\V^n_\w(\F)\) invariant under \(\lambda \star v = \lambda^{q_j} v\). Any proper subspace \(W \subset V_j\) is not invariant unless \(W = \{0\}\).
\end{cor}

\begin{proof}
For \(v = x e_j \in V_j\), we have \(\lambda \star v = \lambda^{q_j} x e_j \in V_j\), so \(V_j\) is invariant. Suppose \(U \supset V_j\) is invariant and contains \(u = x_j e_j + x_k e_k\), \(k \neq j\). By Lemma \ref{grad-inv}, \(U\) is spanned by homogeneous vectors. Since \(\lambda \star u = \lambda^{q_j} x_j e_j + \lambda^{q_k} x_k e_k \in U\) and \(q_j \neq q_k\), choosing \(\lambda\) with \(\lambda^{q_j} \neq \lambda^{q_k}\) implies \(e_j, e_k \in U\), so \(U = \V^n_\w(\F)\). Thus, \(V_j\) is maximal. Since \(V_j\) is one-dimensional, any proper \(W \subset V_j\) is \(\{0\}\), which is invariant but trivial.
\end{proof}

 
\section{Graded Neural Networks}
\label{sec-3}

Graded Neural Networks, introduced in \cite{sh-89} and building on \cite{2024-2}, embed algebraic grading to prioritize features via attention mechanisms and loss weighting, forming the foundation for the Graded Transformer; see \cite{sh-89} for details.
 
Let $\V_\w^n(\F) $ be as in the previous section with 
 $n \geq 1$,  and   $\w = (q_0, \ldots, q_{n-1}) \in \Q^n$,  which  defines the \textbf{grades}, with $\gr(x_i) = q_i$.

A \textbf{graded neuron} on $\V_\w (\F)$ is typically defined as an additive map $\alpha_\w: \V_\w^n \to \F$ such that 
\[
\alpha_\w(\x) = \sum_{i=0}^{n-1} w_i^{q_i} x_i + b,
\]
where $w_i \in \F$ are \textbf{neural weights}, and $b \in \F$ is the \textbf{bias}. For $b = 0$, 
\[
\alpha_\w(\lambda \star \x) = \sum (\lambda w_i)^{q_i} x_i = \lambda \sum w_i' x_i
\]
for ($w_i' = w_i^{q_i}$), approximating a graded linear map of degree 1, as in  \cref{sec-2}.

Alternatively, a \textbf{multiplicative graded neuron} can be defined as $\beta_\w: \V_\w^n \to \F$ such that 
\[
\beta_\w(\x) = \prod_{i=0}^{n-1} (w_i x_i)^{q_i} + b,
\]
capturing multiplicative interactions among graded features. For $b = 0$, 
\[
\beta_\w(\lambda \star \x) = \prod (\lambda^{q_i} w_i x_i)^{q_i} = \lambda^{\sum q_i^2} \prod (w_i x_i)^{q_i},
\]
reflecting a higher-degree graded map, enhancing expressivity for nonlinear relationships.

\begin{rem}
We denote neural weights by $w$ and grades by $q_i$, with exponents $w_i^{q_i}$ (or $(w_i x_i)^{q_i}$ in multiplicative neurons) reflecting grading, while $w_i$ are parameters optimized during training. In mathematics, \emph{graded spaces} are sometimes referred to as \emph{weighted spaces}, which may cause confusion.
\end{rem}

A \textbf{graded network layer} is:
\[
\begin{aligned}
\phi: \V_\w^n(\F) &\to \V_\w^n(\F) \\
\x &\to g(\W \x + \bb),
\end{aligned}
\]
where 
\[
\W = [w_{j,i}^{q_i}] \in \F^{n \times n}, \quad \bb = [b_0, \ldots, b_{n-1}]^t \in \F^n,
\]
and \(\phi\) preserves grading, with \(\gr(y_j) = q_j\). 

Layers using multiplicative neurons, \(\phi(\x) = g(\prod_i ((\W \x)_i)^{q_i} + \bb)\), are also possible but increase computational complexity; see \cite{2024-2}. Hybrid layers (additive/multiplicative) offer flexibility for diverse applications.

A \textbf{graded neural network} (GNN) is a composition of multiple layers given as 
\[
\hat{\y} = \phi_m \circ \cdots \circ \phi_1 (\x),
\]
where each layer 
\[
\phi_l(\x) = g_l(\W^l \x + \bb^l)
\]
applies a transformation defined by the matrix of neural weights \(\W^l = [w_{j,i}^{q_i}]\), producing outputs \(\hat{\y}\) and true values \(\y\) in \(\V_\w^n\) with grades \(\gr(\hat{y}_i) = q_i\).

\subsection{ReLU Activation}
In classical neural networks, the rectified linear unit (ReLU) activation, defined as $\relu(x) = \max \{ 0, x \}$, applies a simple thresholding to promote sparsity and efficiency. However, for graded neural networks a direct application of this ReLU ignores the grading’s intrinsic scaling. To adapt to this structure, we define a \emph{graded ReLU} that adjusts nonlinearity by grade.
For $\x \in \V_\w^n$, the graded ReLU is:
$$\relu_i(x_i) = \max \{ 0, |x_i|^{1/q_i} \},$$
and
$$\relu(\x) = (\relu_0(x_0), \ldots, \relu_{n-1}(x_{n-1})).$$
Unlike the classical $\max \{ 0, x_i \}$, which treats all coordinates uniformly, this version scales each $x_i$ by $1/q_i$, reflecting the graded action.

For $\lambda \star \x$ we have
\[
\begin{aligned}
\relu_i (\lambda^{q_i} x_i) 
& = \max \{ 0, |\lambda^{q_i} x_i|^{1/q_i} \} = \max \{ 0, |\lambda| |x_i|^{1/q_i} \} \\
& = |\lambda| \max \{ 0, |x_i|^{1/q_i} \},  = |\lambda| \relu_i(x_i), 
\end{aligned}
\]
for  $\lambda \in \F$.
Thus, 
\[
\relu(\lambda \star \x) = |\lambda| \cdot \relu(\x),
\]
 unlike classical ReLU, where $\relu(\lambda \x) = \lambda \relu(\x)$, 
  for $\lambda > 0$ assumes degree 1 homogeneity.

\begin{rem}
Unlike classical ReLU, which outputs 0 for negative inputs, the graded ReLU gives 
\[
\relu_i(-x_i) = |x_i|^{1/q_i},
\]
 reducing sparsity and potentially affecting GNN training efficiency. An alternative, $\max \{ 0, |x_i|^{1/q_i} \operatorname{sgn}(x_i) \}$, restores thresholding while maintaining graded scaling.
\end{rem}

\subsection{Exponential Graded Activation}
An alternative activation is the \textbf{exponential graded activation} defined as 
\[
\exp_i(x_i) = \exp\left( \frac{x_i}{q_i} \right) - 1,
\]
and 
\begin{equation}
\exp(\x) = (\exp_0(x_0), \ldots, \exp_{n-1}(x_{n-1})).
\end{equation}
This activation mitigates numerical instability for large \(q_i\) by inversely scaling inputs, ensuring smoother gradients and more gradual growth than \(\relu_i\) for large positive inputs. For \(\lambda \star \x\), 
\[
\exp_i(\lambda^{q_i} x_i) = \exp\left( \frac{\lambda^{q_i} x_i}{q_i} \right) - 1.
\]
Thus, 
\[
\exp(\lambda \star \x) = \left( \exp\left( \frac{\lambda^{q_0} x_0}{q_0} \right) - 1, \ldots, \exp\left( \frac{\lambda^{q_{n-1}} x_{n-1}}{q_{n-1}} \right) - 1 \right),
\]
reflecting the graded scaling.

\subsection{Graded Loss Functions}
In classical neural networks, loss functions like MSE 
\[
L = \frac{1}{n} \sum_{i=0}^{n-1} (y_i - \hat{y}_i)^2
\]
 treat coordinates equally. In contrast, graded losses for $\V_\w^n(\F)$ with grades $\gr(x_i) = q_i$ prioritize errors via $\w = (q_0, \ldots, q_{n-1})$, enhancing efficiency in structured tasks like algebraic geometry or NLP.  This prioritization is crucial for the Graded Transformer, where graded losses weight errors by algebraic degree to reduce sample complexity.

We define norm-based losses for regression (emphasizing error magnitude) and classification losses for categorical tasks, leveraging $q_i$ for graded alignment. See \cite{sh-89} for details. These losses weight errors by algebraic degree, enhancing efficiency in structured data tasks.

The \textbf{graded MSE} on $\V_\w^n$ is:
\begin{equation}
\Lmse(\y, \hat{\y}) = \frac{1}{n} \sum_{i=0}^{n-1} q_i (y_i - \hat{y}_i)^2,
\end{equation}
where $\y, \hat{\y} \in \V_\w^n$ are true and predicted values, and $q_i$ amplifies errors for higher-graded coordinates  assuming $q_i > 0$.
This scales with grading: for $\lambda \star (\y - \hat{\y}) = (\lambda^{q_i} (y_i - \hat{y}_i))$,
\[
\Lmse(\lambda \star \y, \lambda \star \hat{\y}) = \frac{1}{n} \sum q_i \lambda^{2 q_i} (y_i - \hat{y}_i)^2,
\]
reflecting $\V_\w^n$’s geometry. Alternatively, the \textbf{graded norm loss} uses the graded Euclidean norm from \cref{sec-2}:
\begin{equation}
\Lnorm(\y, \hat{\y}) = \norm{\y - \hat{\y}}_\w^2 = \sum_{i=0}^{n-1} q_i |y_i - \hat{y}_i|^2,
\end{equation}
omitting the $\frac 1 n$ normalization for direct alignment with $\norm{\cdot}_\w$.

The \textbf{homogeneous loss} leverages the homogeneous norm from \cref{sec-2}. For $\V_\w^n$ with distinct grades $d_1, \ldots, d_r$ in $\w$, let $(\y - \hat{\y})_{d_j}$ denote the components corresponding to grade $d_j$. Then:
\[
\Lhom(\y, \hat{\y}) = \norm{\y - \hat{\y}}^2 = \left( \sum_{j=1}^r \norm{(\y - \hat{\y})_{d_j}}_{d_j}^{2 (r - j+1)} \right)^{\frac 1 r},
\]
where $r$ is the number of distinct grades, and $\norm{\cdot}_{d_j}$ is the Euclidean norm on the subspace of grade $d_j$. 

For example, for $\V_{(2,3)}$ with $d_1 = 2$, $d_2 = 3$, and $r = 2$
\[
\Lhom(\y, \hat{\y}) = \left( \norm{(\y - \hat{\y})_2}_2^4 + \norm{(\y - \hat{\y})_3}_3^2 \right)^{1/2}.
\]
This emphasizes higher-graded errors   over lower-graded ones.

For classification, the \textbf{graded cross-entropy} is:
\[
\Lce(\y, \hat{\y}) = - \sum_{i=0}^{n-1} q_i y_i \log(\hat{y}_i),
\]
assuming $\hat{y}_i$ are probabilities (e.g., via softmax), weighting log-losses by $q_i$. The \textbf{max-graded loss} is:
\[
\Lmax(\y, \hat{\y}) = \norm{\y - \hat{\y}}_{\text{max}}^2 = \left( \max_{i} \{ q_i^{1/2} |y_i - \hat{y}_i| \} \right)^2,
\]
focusing on the largest grade-adjusted error, akin to $L_\infty$.

\section{Standard Transformers}
\label{sec-4}

The transformer architecture has become a cornerstone of modern sequence modeling, excelling in tasks from natural language processing to time-series analysis \cite{vaswani2017attention,wen2023transformers}. However, its limitations in capturing hierarchical data structures, such as nested polynomial degrees in algebraic geometry or multi-level protein folding in biological sequences, motivate the development of the graded transformer, which builds on our prior work in Graded Neural Networks \cite{2024-2, sh-89}. This section outlines the standard transformer’s components—input embedding, encoder, decoder, and autoregressive generation—providing a mathematically rigorous foundation for understanding its strengths and shortcomings, setting the stage for the graded transformer’s advancements in \cref{sec-5} and \cref{sec-6}.

\subsection{Transformer Definition}

Let \( \Voc = \{1, 2, \ldots, |\Voc|\} \) be a finite vocabulary of tokens. For \( d \in \mathbb{N} \), let \( \mathbb{R}^d \) be the embedding space for token representations. Define \( \Voc^n \) as the set of sequences of length \( n \) over \( \Voc \). 
A sequence \( (t_1, \ldots, t_n) \in \Voc^n \) is represented as a matrix in \( \Mat_{n,d}(\mathbb{R}) \) after embedding and positional encoding, as defined in subsequent subsections.

\index{transformer}
\index{learnable parameters \( \theta \)}

A \textbf{transformer} is a function
\begin{equation}\label{def:transformer}
\cT_\theta : \cS_{\text{in}} \to \cS_{\text{out}},
\end{equation}
where \( \cS_{\text{in}} = \bigcup_{n=1}^{n_{\max}} \Voc^n \) and \( \cS_{\text{out}} = \bigcup_{m=1}^{m_{\max}} \Voc^m \), with \( n_{\max} \) and \( m_{\max} \) being implementation-specific maximum lengths. It is parameterized by a collection of \textbf{learnable parameters \( \theta \)}, mapping an input sequence \( t = (t_1, \ldots, t_n) \in \Voc^n \) to an output sequence \( s = (s_1, \ldots, s_m) \in \Voc^m \), with \( m \leq m_{\max} \) determined by the generation process. The parameters \( \theta \) include embedding matrices, attention weights, feed-forward weights, and normalization parameters, specified in each component.

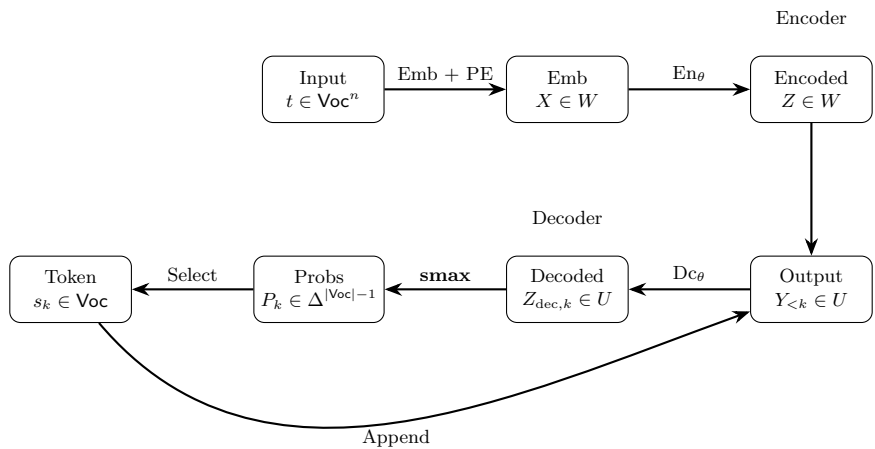
\begin{figure}[H]
    \centering
    \begin{tikzpicture}[
        scale=0.8, transform shape,
        node distance=0.8cm and 2cm,
        box/.style={rectangle, draw, rounded corners, minimum height=1.1cm, minimum width=2cm, align=center, font=\small},
        arrow/.style={-Stealth, thick}
    ]
        \node[box] (input) {Input\\ \( t \in \Voc^n \)};
        \node[box, right=of input] (X) {Emb\\ \( X \in W \)};
        \node[box, right=of X] (Z) {Encoded\\ \( Z \in W \)};
        \node[box, below=2.2cm of Z] (Y) {Output\\ \( Y_{<k} \in U \)};
        \node[box, left=of Y] (Zdec) {Decoded\\ \( Z_{\text{dec},k} \in U \)};
        \node[box, left=of Zdec] (Pt) {Probs\\ \( P_k \in \Delta^{|\Voc|-1} \)};
        \node[box, left=of Pt] (st) {Token\\ \( s_k \in \Voc \)};

        \draw[arrow] (input) -- node[above, font=\small] {Emb + PE } (X);
        \draw[arrow] (X) -- node[above, font=\small] {\(\En_\theta\)} (Z);
        \draw[arrow] (Z) -- (Y);
        \draw[arrow] (Y) -- node[above, font=\small] {\(\Dc_\theta\)} (Zdec);
        \draw[arrow] (Zdec) -- node[above, font=\small] {\(\bsf\)} (Pt);
        \draw[arrow] (Pt) -- node[above, font=\small] {Select} (st);
        \draw[arrow] (st) to[out=-50, in=200] node[below, font=\small] {Append} (Y);

        \node[above=0.4cm of Z, font=\small] {Encoder};
        \node[above=0.4cm of Zdec, font=\small] {Decoder};
    \end{tikzpicture}
    \caption{Transformer architecture mapping input sequence \( t \in \Voc^n \) to output sequence \( s \in \Voc^m \), where \( W = \Mat_{n,d}(\mathbb{R}) \) represents embedded inputs and \( U = \Mat_{k,d}(\mathbb{R}) \) represents partial outputs at step \( k \), processed through encoder and decoder stages. The feedback loop enables autoregressive generation.}
    \label{fig:transformer}
\end{figure}

The transformer comprises three main stages: input embedding with positional encoding, an encoder, and an autoregressive decoder. The architecture is illustrated in \cref{fig:transformer}.

\subsection{Input Embedding and Positional Encoding}
\label{subsubsec:input_representation}

To process discrete token sequences in \( \Voc^n \) through the transformer’s neural architecture, input embeddings map tokens to continuous vectors, while positional encodings preserve sequence order, addressing the permutation invariance of attention mechanisms.

Define the embedding matrix \( \mathbf{W}_e \in \Mat_{|\Voc|,d}(\mathbb{R}) \). For a token \( t_i \in \Voc \), its embedding is \( \x_i = \mathbf{W}_e \cdot \mathbf{e}_{t_i} \in \mathbb{R}^d \), where \( \mathbf{e}_{t_i} \in \mathbb{R}^{|\Voc|} \) is the one-hot vector with a 1 at index \( t_i \) and zeros elsewhere.

\index{embedding}
\index{positional encoding}

To incorporate positional information, define the \textbf{positional encoding} function
$
\pe : \{1, \ldots, n_{\max}\} \to \mathbb{R}^d,
$  
given by:
\begin{equation}
\pe(i)_k =
\begin{cases}
\sin\left( \frac{i}{10000^{k / d}} \right) & \text{if } k \text{ is even}, \\
\cos\left( \frac{i}{10000^{(k-1) / d}} \right) & \text{if } k \text{ is odd},
\end{cases}
\end{equation}
for \( k = 0, \ldots, d-1 \). For an input sequence \( t = (t_1, \ldots, t_n) \in \Voc^n \), the embedded input matrix is:
\begin{equation}
X = [\x_1 + \pe(1), \ldots, \x_n + \pe(n)]^T \in \Mat_{n,d}(\mathbb{R}),
\end{equation}
where the rows correspond to \( \x_i + \pe(i) \). These fixed encodings, while effective for sequential data, lack structural priors for hierarchical data, a limitation addressed by the graded transformer in \cref{sec-5}.

\subsection{Encoder}
\label{subsubsec:encoder}

The encoder transforms the embedded input matrix \( X \) into contextualized representations, capturing global sequence dependencies through stacked layers of attention and transformation.
The encoder is a function:
\begin{equation}
\En_\theta : \Mat_{n,d}(\mathbb{R}) \to \Mat_{n,d}(\mathbb{R}),
\end{equation}
transforming \( X \) into a contextualized representation \( Z = \En_\theta(X) \). It consists of \textbf{ \( N \) layers}, each applying multi-head self-attention, a feed-forward network, residual connections, and layer normalization. While effective for sequential data, this structure struggles with hierarchical patterns, a limitation addressed by the graded transformer in \cref{sec-5}.

\index{softmax function}

The \textbf{softmax function}   $ \bsf : \Mat_{m,n}(\mathbb{R}) 		\to \Mat_{m,n}(\mathbb{R}) $ 
normalizes attention scores into probabilities:
\[
\begin{split}
\bsf(M)_{i,j} 				&	= \frac{\exp(m_{i,j})}{\sum_{k=1}^n \exp(m_{i,k})},
\end{split}
\]
for a matrix \( M = [m_{i,j}] \in \Mat_{m,n}(\mathbb{R}) \), where the image of $\bsf$ is 
\[
 \left\{ P \in \Mat_{m,n}(\mathbb{R}) \mid p_{i,j} \geq 0, \sum_{j=1}^n p_{i,j} = 1 \text{ for all } i \right\},
\]

\subsubsection{Self-Attention Mechanism}

Self-attention computes weighted representations of input tokens by allowing each token to attend to all others in the sequence, capturing dependencies dynamically regardless of distance, unlike fixed convolutional filters. In its single-head form, it projects the input into query (\( Q \)), key (\( K \)), and value (\( V \)) matrices via learned linear transformations (parameterized by weight matrices $\W_Q$, $\W_K$, $\W_V$), optimized during training to emphasize task-relevant token interactions. Multi-head attention extends this by running multiple parallel attention heads, each learning different aspects of relationships (e.g., syntactic vs. semantic), and concatenating their outputs for richer contextualization. The notation \( V \) for the value matrix, standard in transformer literature, represents features aggregated from matched keys, though care is needed to avoid conflicts with uses of \( V \) for vector spaces or volumes in later algebraic contexts (e.g., \cref{sec-5}).

\index{self-attention}

First, consider single-head self-attention for layer \( k \). Define parameter matrices $\W_{Q,k}$, $\W_{K,k}$, $\W_{V,k} \in \Mat_{d,d}(\R)$. For \( X \in \Mat_{n,d}(\R) \):
\begin{equation}
Q = X \W_{Q,k}, \; K = X \W_{K,k}, \; V = X \W_{V,k}. 
\end{equation}
The \textbf{attention matrix}  is:
\begin{equation}
A_k(X) = \bsf\left( \frac{Q K^T}{\sqrt{d}} \right) V \in \Mat_{n,d}(\R).
\end{equation}
Now, extend to multi-head self-attention. Let \( h \in \N \) be the number of attention heads, and \( d_k = d / h \in \N \). For layer \( k = 1, \ldots, N \), head \( i = 1, \ldots, h \), define parameter matrices $\W_{Q,k,i}$, $\W_{K,k,i}$, $\W_{V,k,i} \in \Mat_{d,d_k}(\R)$. For \( X \in \Mat_{n,d}(\R) \):
\begin{equation}
Q_i = X \W_{Q,k,i}, \;  K_i = X \W_{K,k,i}, \; V_i = X \W_{V,k,i}.   
\end{equation}
The \textbf{attention for head \( i \)} is:
\begin{equation}
A_{k,i}(X) = \bsf\left( \frac{Q_i K_i^T}{\sqrt{d_k}} \right) V_i \in \Mat_{n,d_k}(\R).
\end{equation}
The \textbf{multi-head attention} is:
\begin{equation}
\Mh_k(X) = \Ct(A_{k,1}(X), \ldots, A_{k,h}(X)) \W_{O,k} \in \Mat_{n,d}(\R),
\end{equation}
where $\Ct$ concatenates head outputs along the feature dimension, and $\W_{O,k} \in \Mat_{h d_k, d}(\R)$.

\begin{rem}
Algebraically, multi-head attention induces a representation of the symmetric group via permutation equivariance (\cref{prop:perm_equiv}), treating sequences as multisets. However, this ignores graded hierarchies, such as polynomial rings where degrees require differential weighting. The LGT/EGT extend this to graded modules with degree-aware linear or exponential biases, as detailed in \cref{sec-5}.
\end{rem}

\subsubsection{Properties of the Attention Matrix}

A matrix $ M \in \Mat_{n,n}(\R) $ is \textbf{row-stochastic} if all entries are non-negative and each row sums to 1, i.e., $ M_{j,\cdot} \mathbf{1} = 1 $ for all rows $ j $, where $ \mathbf{1} $ is the all-ones vector. Such matrices represent probability distributions over columns.

A symmetric matrix $ M \in \Mat_{n,n}(\R) $ is \textbf{positive-semidefinite (PSD)} if all its eigenvalues are non-negative, or equivalently, if $ \x^T M \x \geq 0 $ for all vectors $ \x \in \R^n $.

The \textbf{Gram matrix} of vectors $ \{\mathbf{v}_1, \dots, \mathbf{v}_n\} \subset \R^{d_k} $ is $ G_{j,l} = \langle \mathbf{v}_j, \mathbf{v}_l \rangle $, which is always PSD by construction.

The \textbf{Gram kernel sense} refers to interpreting $ S $ as a kernel matrix in kernel methods, where the dot-product kernel $ k(\x, \mathbf{y}) = \langle \x, \mathbf{y} \rangle / \sqrt{d_k} $ induces a PSD Gram matrix if the kernel is positive definite (satisfying Mercer's theorem for valid reproducing kernel Hilbert spaces).

\begin{prop}
The attention matrix 
\[
 A_{k,i} = \bsf(Q_i K_i^T / \sqrt{d_k}) 
 \]
  for head $ i $ in layer $ k $ is row-stochastic. Additionally, the pre-softmax matrix 
  \[
  S = Q_i K_i^T / \sqrt{d_k} 
  \]
   is positive-semidefinite when $ Q_i = K_i $ and the projections are orthogonal, with eigenvalues bounded by $ O(n) $, where $ n $ is the sequence length.
\end{prop}
 
\begin{proof}
The row-stochastic property follows from the softmax function: for each row $ j $ of $ S = Q_i K_i^T / \sqrt{d_k} $, 
\[
A_{k,i,j,l} = \frac {\exp(S_{j,l}) } { \sum_{m=1}^n \exp(S_{j,m}) }   \geq 0,
\]
 and the row sums to 1 by construction.

For the positive-semidefinite  property of $ S $, assume $ Q_i = K_i $ (standard in self-attention with shared projections), making $ S $ symmetric. Then, 
\[
 S_{j,l} = \langle \w_j, \mathbf{k}_l \rangle / \sqrt{d_k} = \langle \w_j, \w_l \rangle / \sqrt{d_k},
 \]
  which is a scaled Gram matrix of the rows $ \{\w_1, \dots, \w_n\} $. Gram matrices are positive-semidefinite, as $ S = (1/\sqrt{d_k}) P^T P $, where $P$ has rows $ \w_j $, and thus eigenvalues are non-negative.

Under orthogonality (i.e., $ \W_{Q,k,i}^T \W_{Q,k,i} = I $, preserving norms), if rows of $ Q_i $ are unit-normalized (e.g., via prior layer normalization), the diagonals $ S_{j,j} = 1 / \sqrt{d_k} $, so the trace 
\[
tr (S) = n / \sqrt{d_k} = O(n / \sqrt{d_k}).
 \]
  Since $S$ is positive-semidefinite, eigenvalues $\lambda$  satisfy 
  \[
  0 \leq \lambda \leq  tr(S) = O(n / \sqrt{d_k}),
  \]
   but in practice with $d_k$ fixed (e.g., 64), it's $O(n)$. 
For the softmax-applied $A_{k,i}$, as a row-stochastic matrix, its eigenvalues lie in the unit disk $|\lambda|\leq 1$  by the Perron-Frobenius theorem, with $\lambda =1$  always an eigenvalue, and others bounded below 1 approximately under the assumptions.

\end{proof}

\begin{rem}
This analysis aligns with the kernel interpretation of self-attention, where \( S \) acts as a dot-product kernel matrix, enabling efficient computation but treating tokens uniformly \cite{tsai2019transformer}. This limits applicability to graded hierarchies, unlike the LGT/EGT’s degree-aware kernels in \cref{sec-5}. Eigenvalue bounds ensure training stability, as analyzed in transformer dynamics \cite{geshkovski2023mathematical}.
\end{rem}

\subsubsection{Feed-Forward Network}

The feed-forward network applies non-linear transformations per token to enhance the encoder’s representational capacity.

\begin{df}[Feed-Forward Network]
The feed-forward network:
\begin{equation}
\fnn_k : \Mat_{n,d}(\R) \to \Mat_{n,d}(\R),
\end{equation}
is applied row-wise. For a row \( \mathbf{z} \in \R^d \):
\begin{equation}
\fnn_k(\mathbf{z}) = \W_{2,k} \cdot \relu(\W_{1,k} \mathbf{z} + \mathbf{b}_{1,k}) + \mathbf{b}_{2,k},
\end{equation}
where $\W_{1,k} \in \Mat_{d,d_f}(\R)$, $\W_{2,k} \in \Mat_{d_f,d}(\R)$, $\mathbf{b}_{1,k} \in \R^{d_f}$, $\mathbf{b}_{2,k} \in \R^d$, $d_f \in \N$.
\end{df}

\begin{rem}
The feed-forward network introduces nonlinearity and expands the dimensionality to $d_f > d$, allowing richer representations. In graded transformers, this can be adapted to respect grading, as detailed in \cref{sec-5}.
\end{rem}

\begin{lem}[Feed-Forward Lipschitz Continuity]
The feed-forward network $\fnn_k$ is Lipschitz continuous with constant bounded by $\|W_{1,k}\|_2 \|  \, W_{2,k}\|_2$, assuming ReLU activation.
\end{lem}

\begin{proof}
ReLU is 1-Lipschitz ($|\relu(a) - \relu(b)| \leq |a - b|$), and linear transformations have Lipschitz constants equal to their spectral norms. By the chain rule, the composition's constant is the product of the individual constants.
\end{proof}

\subsubsection{Layer Normalization}

Layer normalization stabilizes training by normalizing token features, enabling deeper networks and smoother optimization.

\begin{df}[Layer Normalization]
Layer normalization:
\begin{equation}
\Ln : \Mat_{n,d}(\R) \to \Mat_{n,d}(\R),
\end{equation}
is applied row-wise. For a row \( \mathbf{z} \in \R^d \):
\begin{equation}
\Ln(\mathbf{z}) = \frac{\mathbf{z} - \mu}{\sqrt{\sigma^2 + \epsilon}} \cdot \boldsymbol{\gamma} + \boldsymbol{\beta},
\end{equation}
where \( \mu = \frac{1}{d} \sum_{j=1}^d z_j \), \( \sigma^2 = \frac{1}{d} \sum_{j=1}^d (z_j - \mu)^2 \), \( \epsilon > 0 \), and \( \boldsymbol{\gamma}, \boldsymbol{\beta} \in \R^d \).
\end{df}

\begin{lem}
Post-normalization, rows approximately satisfy 
\[
 \|\Ln(\mathbf{z}) - \boldsymbol{\beta}\|_{\boldsymbol{\gamma}} \approx \sqrt{d} 
 \]
  in high dimensions (scaled \(\ell^2\)-norm with \(\boldsymbol{\gamma}\)). 
\end{lem}

\begin{proof}
After normalization, \( (\mathbf{z} - \mu) / \sqrt{\sigma^2 + \epsilon} \) has mean 0 and variance 1 across features, with \(\ell^2\)-norm approximately \(\sqrt{d}\) in high dimensions by the law of large numbers. Element-wise multiplication by \( \boldsymbol{\gamma} \) scales the norm, and the shift by \( \boldsymbol{\beta} \) centers it, resembling a projection onto \(\mathbb{S}^{d-1}\) scaled by \(\sqrt{d}\) when \( \boldsymbol{\gamma} = \mathbf{1} \), \( \boldsymbol{\beta} = \mathbf{0} \).
\end{proof}

\begin{rem}
This resembles a projection onto $\mathbb{S}^{d-1}$ scaled by $\sqrt{d}$ when $ \boldsymbol{\gamma} = \mathbf{1} $, $ \boldsymbol{\beta} = \mathbf{0} $, analogous to the structure of graded vector spaces in LGT/EGT (\cref{sec-5}), where normalization aligns with hierarchical scaling.
\end{rem}

\begin{ex}
For a token embedding \(\mathbf{z} = [0.5, -0.2, 0.3]\) with \(d=3\), if \(\mu = 0.2\), \(\sigma^2 \approx 0.11\), \(\epsilon = 10^{-5}\), \(\boldsymbol{\gamma} = \mathbf{1}\), \(\boldsymbol{\beta} = \mathbf{0}\), then \(\Ln(\mathbf{z}) \approx [0.95, -0.38, 0.57]\), with \(\|\Ln(\mathbf{z})\|_2 \approx \sqrt{3}\), stabilizing gradients for attention layers.
\end{ex}

\subsubsection{Encoder Layer Composition}

The encoder layer integrates multi-head self-attention, feed-forward transformations, residual connections, and layer normalization to produce contextualized token representations while preserving gradient flow.

The \( k \)-th encoder layer is:
\begin{equation}
\EnL_k(X) = \Ln( X' + \fnn_k(X') ), \quad X' = \Ln( X + \Mh_k(X) ),
\end{equation}
where \( \Mh_k \) is multi-head self-attention (\cref{subsubsec:encoder}), \( \fnn_k \) is the feed-forward network, and \( \Ln \) is layer normalization. The encoder, with \( N \) layers, is the composition:
\begin{equation}
\En_\theta(X) = \EnL_N \circ \EnL_{N-1} \circ \cdots \circ \EnL_1(X).
\end{equation}

\begin{rem}
In graded transformers, this structure is adapted by applying grading transformations to prioritize hierarchical features, enhancing efficiency for structured data (cf. \cref{sec-5}).
\end{rem}

\subsection{Limitations for Hierarchical Data}

Transformers require extensive data to learn hierarchies (e.g., syntactic trees or polynomial gradings), as attention lacks priors—provably \(O(e^L)\) sample complexity for depth \(L\) via VC bounds \cite{yun2019transformers}.

Recent mathematical analyses model transformers as particle systems evolving on spheres, where token representations (particles) cluster over time to capture dependencies. However, this clustering is slow in high-dimensional or hierarchical settings, as quantified in the following result from \cite{geshkovski2023mathematical}.

\begin{thm} 
In standard transformers modeled as particle systems on spheres, tokens cluster at an exponential rate \(O(e^{-\lambda t})\) with \(\lambda = O(1/(n e^{2\beta}))\), implying slow convergence (near \(O(1/t)\)) for large \(\beta\) in high-dimensional hierarchical settings.
\end{thm}
 
Here, \(t\) represents training time (or iterations), \(n\) is sequence length, and \(\beta\) is an inverse temperature parameter controlling clustering strength. For hierarchical data with large effective \(\beta\) (e.g., deep syntactic trees or multi-scale physics simulations), convergence slows dramatically, as the rate \(\lambda\) diminishes exponentially. This underscores transformers' inefficiency without structural priors.


\subsection{Decoder}
\label{subsubsec:decoder}

The decoder generates output sequences by integrating encoder outputs with causal self-attention and cross-attention. It is a function:
\begin{equation}
\Dc_\theta : \Mat_{t,d}(\R) \times \Mat_{n,d}(\R) \to \Mat_{t,d}(\R),
\end{equation}
taking \( Y_{<t} \in \Mat_{t,d}(\R) \) (current output sequence) and \( Z \in \Mat_{n,d}(\R) \) (encoder output) to produce \( Y_{\text{dec}} = \Dc_\theta(Y_{<t}, Z) \). It consists of \( N \) layers, each with masked self-attention, cross-attention, and a feed-forward network.

\begin{df}[Masked Multi-Head Self-Attention]
For \( Y \in \Mat_{t,d}(\R) \), let 
\begin{equation}
Q_i = Y \W_{Q,l,i}, \; K_i = Y \W_{K,l,i}, \; V_i = Y \W_{V,l,i} \in \Mat_{t,d_k}(\R),
\end{equation}
with \( \W_{Q,l,i}, \W_{K,l,i}, \) and \( \W_{V,l,i} \in \Mat_{d,d_k}(\R) \). The \textbf{masked attention}  is defined as 
\begin{equation}
A_{l,i}(Y) = \bsf\left( \frac{Q_i K_i^T}{\sqrt{d_k}} \cdot M \right) V_i,
\end{equation}
where \( M \in \Mat_{t,t}(\R) \), \( M_{i,j} = 1 \) if \( j \leq i \), and \( M_{i,j} = -\infty \) if \( j > i \) (implemented via masking before softmax). 
Then the \textbf{masked multi-head self-attention} is defined as 
\begin{equation}
\Mmh_l(Y) = \Ct(A_{l,1}(Y), \ldots, A_{l,h}(Y)) \W_{O,l} \in \Mat_{t,d}(\R).
\end{equation}
\end{df}

Next we define the \textbf{cross-attention} \( A_{l,i}(Y, Z) \) as follows, using projections to compute the output:

\begin{df}[Cross-Attention]
For \( Y \in \Mat_{t,d}(\R) \), \( Z \in \Mat_{n,d}(\R) \), the cross-attention is defined using the following projections to compute the output:
\begin{small}
\begin{equation}
\begin{split}
Q_i 		&	= Y \W_{Q,l,i} \in \Mat_{t,d_k}(\R), \\
K_i 		&	= Z \W_{K,l,i},  \\
V_i 		&	= Z \W_{V,l,i} \in \Mat_{n,d_k}(\R), \\
A_{l,i}(Y, Z) 	&	= \bsf\left( \frac{Q_i K_i^T}{\sqrt{d_k}} \right) V_i \in \Mat_{t,d_k}(\R), \\
\Ca_l(Y, Z) 	&	= \Ct(A_{l,1}(Y, Z), \ldots, A_{l,h}(Y, Z)) \W_{O,l}.   
\end{split}
\end{equation}
\end{small}
\end{df}
The \textbf{\( l \)-th decoder} layer is defined as 
\begin{equation}
\Dcl_l(Y, Z) = \Ln( Y_l^{(2)} + \fnn_l(Y_l^{(2)}) ),
\end{equation}
where
\[
\begin{split}
Y_l^{(1)}  & = \Ln( Y + \Mmh_l(Y) ), \\
 Y_l^{(2)}  & = \Ln( Y_l^{(1)} + \Ca_l(Y_l^{(1)}, Z) ).
\end{split} 
\]
The \textbf{decoder} is defined as 
\begin{equation}
\Dc_\theta(Y, Z) = \Dcl_N \circ \cdots \circ \Dcl_1(Y, Z).
\end{equation}

\begin{rem}
Masking in self-attention ensures causality, preventing future token leakage. Structural priors, like those in algebraic frameworks, can enhance masking by weighting tokens based on their hierarchical importance, as explored in \cref{sec-5}.
\end{rem}

\subsection{Autoregressive Generation}
\label{subsubsec:generation}

The transformer’s ability to generate coherent sequences token by token is a cornerstone of its success in applications like language modeling and sequence translation \cite{vaswani2017attention}. Autoregressive generation leverages the encoder’s contextualized representations to produce output sequences iteratively, a process that excels in capturing sequential dependencies but falters when faced with hierarchical data structures, requiring \(O(e^L)\) samples for depth \(L\) hierarchies. This subsection formalizes the standard transformer’s generation mechanism, providing a clear baseline for the trailblazing advancements introduced by the graded transformer in \cref{sec-6}.

For an input sequence \( t \in \Voc^n \), compute the embedded input matrix \( X \in \Mat_{n,d}(\R) \) as in \cref{subsubsec:input_representation} and the encoder output \( Z = \En_\theta(X) \). Initialize with a start token \( s_0 \in \Voc \), setting the initial output sequence:
\begin{equation}
Y_{<1} = [\y_0 + \pe (1)]^T \in \Mat_{1,d}(\R), \quad \y_0 = \W_e \cdot e_{s_0}.
\end{equation}

Generation proceeds recursively: \(Z_{\text{dec},t} = \Dc_\theta(Y_{<t}, Z)\), \(\z_t = (Z_{\text{dec},t})_{t,:}\), \(s_t = \arg\max \bsf(\W_e^T \z_t)\), terminating at EOS or \(m_{\max}\).

This process produces the output sequence \( \cT_\theta(t) = (s_1, \ldots, s_m) \), enabling flexible sequence generation. However, its reliance on unstructured attention limits its ability to prioritize hierarchical features, a challenge overcome by the graded transformer’s algebraic framework, rooted in fixed grades from \cite{2024-2, sh-89} with an extension to learnable grades (\cref{subsec:gt_training_challenges}), as detailed in \cref{sec-6}.

\begin{lem}\label{thm:transformer}
The transformer \( \cT_\theta : \cS_{\text{in}} \to \cS_{\text{out}} \) is a well-defined function.
\end{lem}

\begin{proof}
Being well-defined follows from determinism of matrix operations and softmax normalization.
\end{proof}

This framework, while pivotal for sequence modeling, lacks structural biases for hierarchical data, a limitation addressed by the graded transformer in \cref{sec-6}, building on \cite{2024-2, sh-89} to incorporate algebraic priors for enhanced efficiency and interpretability.

\begin{lem}[Permutation-Equivariance]
\label{prop:perm_equiv}
The self-attention operator:
\begin{equation}
A_{l,i}(X) = \bsf\left( \frac{Q_i K_i^T}{\sqrt{d_k}} \right) V_i,
\end{equation}
is permutation-equivariant. For any permutation matrix \( P \in \Mat_n(\R) \), if:
$Q_i' = P Q_i$, $K_i' = P K_i$, $V_i' = P V_i$,  then:
\begin{equation}
A_{l,i}(Q_i', K_i', V_i') = P A_{l,i}(Q_i, K_i, V_i).
\end{equation}
\end{lem}

\begin{proof}
Let \( S = \frac{Q_i K_i^T}{\sqrt{d_k}} \). Then we have 
\[
Q_i' K_i'^T = P Q_i K_i^T P^T = P S P^T,
\]
and  $\bsf(P S P^T) = P\cdot  \bsf(S) \cdot   P^T$. 
Thus, 
\begin{equation}
\begin{split}
A_{l,i}(Q_i', K_i', V_i')  & = \bsf(P S P^T) (P V_i) = P \, \bsf(S) \, V_i \\
& = P A_{l,i}(Q_i, K_i, V_i).
\end{split}
\end{equation}
This completes the proof. 
\end{proof}

\begin{prop}[Computational Complexity]
\label{prop:complexity-1}
The complexity of a self-attention head \( A_{l,i}(X) \) is \( O(n^2 d_k) \), and for multi-head attention with \( h \) heads, \( O(n^2 d) \). Quadratic \(n^2\) arises from full pairwise interactions; graded attention reduces to \(O(n d)\) via sparse hierarchies.
\end{prop}

\begin{proof}
For one head: computing \( Q_i, K_i, V_i \) costs \( O(n d d_k) \); \( Q_i K_i^T \) costs \( O(n^2 d_k) \); softmax costs \( O(n^2) \); and \( \bsf(Q_i K_i^T / \sqrt{d_k}) V_i \) costs \( O(n^2 d_k) \). The total is dominated by \( O(n^2 d_k) \). For \( h \) heads, with \( d_k = d/h \), the complexity is \( O(h \cdot n^2 d/h) = O(n^2 d) \). The feed-forward network costs \( O(n d d_f) \).
\end{proof}

The computational cost of attention, while significant, is balanced by careful scaling of attention scores to ensure numerical stability, as detailed below.

\begin{prop}[Scaling Factor]
\label{prop:scaling_factor}
Let \( Q, K \in \R^{n \times d_k} \), where each row \( q_i, k_j \in \R^{d_k} \) consists of independent entries sampled from \( \mathcal{N}(0, 1) \). For the dot-product attention matrix \( S \in \R^{n \times n} \), defined by:
\begin{equation}
S_{i,j} = \frac{q_i \cdot k_j}{\sqrt{d_k}},
\end{equation}
each entry has variance $\mathrm{Var}(S_{i,j}) = 1$. This unit variance prevents saturation in softmax for high \(d\).
\end{prop}

\begin{proof}
Fix rows \( q_i, k_j \in \R^{d_k} \), where \( q_i = (Q_{i,1}, \ldots, Q_{i,d_k}) \), \( k_j = (K_{j,1}, \ldots, K_{j,d_k}) \), with all \( Q_{i,\ell}, K_{j,\ell} \sim \mathcal{N}(0, 1) \), assuming random initialization with independent entries. Then:
\begin{equation}
q_i \cdot k_j = \sum_{\ell=1}^{d_k} Q_{i,\ell} K_{j,\ell}.
\end{equation}
Each product \( Q_{i,\ell} K_{j,\ell} \) has mean zero and variance 1, since the product of two independent \( \mathcal{N}(0, 1) \) variables has variance 1. Since the \( d_k \) terms are independent:
\begin{equation}
\mathrm{Var}(q_i \cdot k_j) = \sum_{\ell=1}^{d_k} \mathrm{Var}(Q_{i,\ell} K_{j,\ell}) = d_k.
\end{equation}
After scaling we have 
$
\mathrm{Var}\left( \frac{q_i \cdot k_j}{\sqrt{d_k}} \right) = \frac{d_k}{d_k} = 1.
$
\end{proof}

\section{Linearly Graded Transformers}\label{sec-5}

The Linearly Graded Transformer (LGT) realizes the core motivation of graded vector spaces \cite{2024-2, sh-89} by treating vectors scaled by weights   as equivalent to their normalized form \([x_0, \dots, x_{d-1}]\), focusing on relative hierarchies without exponential sensitivity. The LGT employs direct linear weights, preserving the algebraic purity of graded spaces for stable and interpretable modeling in domains like algebraic geometry and natural language processing (NLP). This section defines the LGT, proves its mathematical properties, details its architecture, and outlines training strategies, laying the foundation for applications in \cref{sec-7} and \cref{sec-8}. LGT's linear scaling ensures purity in graded modules, contrasting EGT's exponential for deep hierarchies.
We begin by formalizing the LGT architecture, starting with notation and its definition.

Let \((\R^d)^n\) denote the space of sequences \((\x_1, \ldots, \x_n)\), where \(\x_i \in \R^d\). Given a grading tuple \(\w = (q_0, \ldots, q_{d-1})\), define weights \(q_i = f(q_i)\), where \(f: \R \to \R^+\) is a positive function (e.g., \(f(q_i) = |q_i| + 1\)). The \textbf{grading matrix}  is 
\begin{equation}\label{grading-matrix}
\Mlw = \diag(q_0, \ldots, q_{d-1}),
\end{equation}
with linear map \(\Llw: \V \to \V\) given by  \(\Llw(\x) = [q_0 x_0, \ldots, q_{d-1} x_{d-1}]^t \). Define the map 
\begin{equation}
\begin{split}
\phi_{\w}: (\R^d)^n   			&		\to (\R^d)^n, \\
 X = ( \x_1, \ldots, \x_n   )		&		\to 	 ( \Mlw\x_1, \ldots, \Mlw \x_n )
\end{split}
\end{equation}

Let  \(\cT: (\R^d)^n \to (\R^d)^n\) be a given a standard transformer as in \cref{def:transformer}.    Now we are ready for the following definition. 

\begin{df}\label{df:lgt}
The map
\begin{equation}
\begin{split}
\Lgt: (\R^d)^n    &	\to (\R^d)^n, \\
 X 			&	\to  \cT  (\phi_{\w}(X)) = \cT  (  \Mlw\x_1, \ldots, \Mlw \x_n    )
\end{split}
\end{equation}
is called a  \textbf{Linearly Graded Transformer} (LGT). 
\end{df}

\begin{lem}\label{lem:lgt_well_defined}
\(\Lgt: (\R^d)^n \to (\R^d)^n\) is a well-defined map.
\end{lem}

\begin{proof}
The map \(\phi_{\w}\) is well-defined, as \(\Mlw \in \Mat_d(\R)\) is a linear transformation (diagonal with positive entries \(q_i\)), and applying \(\Mlw \x_i \in \R^d\) to each \(\x_i\) yields a sequence \((\Mlw \x_i)_{i=1}^n \in (\R^d)^n\). The standard transformer \(\cT\) is well-defined (\cref{thm:transformer}). Thus, the composition 
\[
\Lgt = \cT \circ \phi_{\w}
\]
 maps \(X \in (\R^d)^n\) to a unique \(Y \in (\R^d)^n\), ensuring that \(\Lgt\)  is well defined.
\end{proof}

This composition embeds grading into the transformer, enabling hierarchical prioritization while preserving computational efficiency.

\begin{figure}[h]
    \centering
    \begin{tikzpicture}[
        scale=0.8, transform shape,
        node distance=0.8cm and 2cm,
        box/.style={rectangle, draw, rounded corners, minimum height=1.1cm, minimum width=2cm, align=center, font=\small},
        arrow/.style={-Stealth, thick}
    ]
        \node[box] (input) {Input\\ \( t \in \Voc^n \)};
        \node[box, right=of input] (X) {Emb\\ \( X \in W \)};
        \node[box, right=of X] (phi) {Grading\\ \( \phi_\w(X) \)};
        \node[box, right=of phi] (Z) {Encoded\\ \( Z \in W \)};
        \node[box, below=2.2cm of Z] (Y) {Output\\ \( Y_{<t} \in U \)};
        \node[box, left=of Y] (Zdec) {Decoded\\ \( Z_{\text{dec},t} \in U \)};
        \node[box, left=of Zdec] (Pt) {Probs\\ \( P_t \in \Delta^{|\Voc|-1} \)};
        \node[box, left=of Pt] (st) {Token\\ \( s_t \in \Voc \)};

        \draw[arrow] (input) -- node[above, font=\small] {Emb + PE } (X);
        \draw[arrow] (X) -- node[above, font=\small] {Lin. Grading} (phi);
        \draw[arrow] (phi) -- node[above, font=\small] {\(\En_\theta\)} (Z);
        \draw[arrow] (Z) -- (Y);
        \draw[arrow] (Y) -- node[above, font=\small] {\(\Dc_\theta\)} (Zdec);
        \draw[arrow] (Zdec) -- node[above, font=\small] {\(\bsf\)} (Pt);
        \draw[arrow] (Pt) -- node[above, font=\small] {Select} (st);
        \draw[arrow] (st) to[out=-50, in=200] node[below, font=\small] {Append} (Y);

        \node[above=0.4cm of Z, font=\small] {Encoder};
        \node[above=0.4cm of Zdec, font=\small] {Decoder};
    \end{tikzpicture}
    \caption{Linearly Graded Transformer (LGT) architecture mapping input sequence \( t \in \Voc^n \) to output sequence \( s \in \Voc^m \), where \( \W = \Mat_{n,d}(\R) \) represents embedded inputs and \( U = \Mat_{t,d}(\R) \) represents partial outputs. The linear grading map \( \phi_\w \) is applied after embedding to prioritize hierarchical features before processing through encoder and decoder stages.}
    \label{fig:lgt_transformer}
\end{figure}
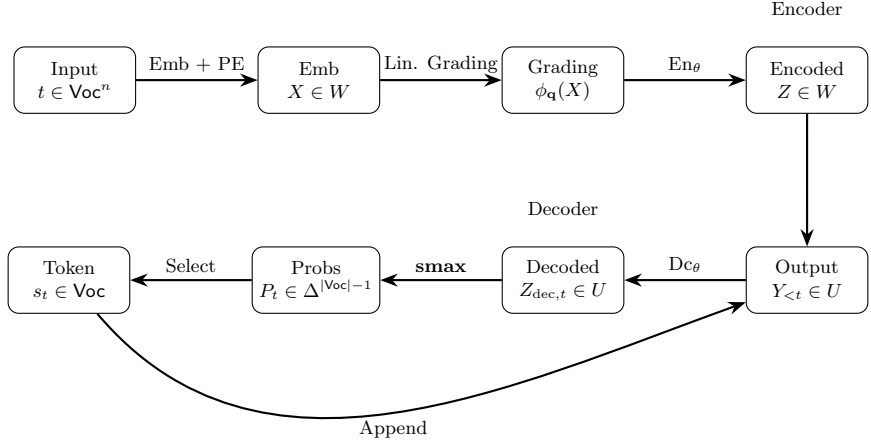

To embed algebraic inductive biases rigorously, we situate the LGT within the category of graded vector spaces with weighted scalar action, extending the framework from \cref{sec-2}. This categorical perspective ensures that the LGT preserves structural properties like equivariance and invariance, crucial for hierarchical data in domains such as algebraic geometry and NLP.

Let \(\mathcal{C}\) be the category of graded vector spaces with weighted scalar action, where objects are graded vector spaces \(\V_\w^n(\R)\) over \(\R\) with decomposition \(\V = \bigoplus_{j=0}^{d-1} V_j\) into weight spaces. The weighted scalar action is \(\lambda \star v = \lambda^{q_j} v\) for \(v \in V_j\) and \(\lambda \in \R^\times\), with grading tuple \(\w = (q_0, \dots, q_{d-1}) \in \Q_{>0}^d\). Morphisms \(f: \V_\w^n(\R) \to \V_{\w'}^m(\R)\) are linear maps that are equivariant (\(f(\lambda \star v) = \lambda \star f(v)\)) and preserve weights (\(f(V_j) \subseteq V'_j\) for compatible \(\w, \w'\)).

For sequences \(X \in (\R^d)^n \cong \V_\w^n(\R)\), the action extends component-wise: \(\lambda \star X = (\lambda \star x_1, \dots, \lambda \star x_n)\). Morphisms \(f\) extend component-wise: \(f(X) = (f(x_1), \dots, f(x_n))\), where \(f\) is a morphism on \(\V_\w(\R)\). Invariant subspaces \(W \subseteq \V_\w^n(\R)\) are those closed under the action (\(\lambda \star W \subseteq W\) for all \(\lambda \in \R^\times\)).

The grading map \(\phi_\w: \V_\w^n(\R) \to \V_\w^n(\R)\) is defined by \(\phi_\w(X) = (M x_1, \dots, M x_n)\), where \(M = \Mlw = \diag(q_0, \dots, q_{d-1})\) and \(q_i = f(q_i) > 0\) for a positive function \(f\).

Since \(M\) is diagonal and acts as a scalar multiple on each weight space \(V_j\) (coordinates in \(V_j\) have the same \(q_j\)), \(M\) commutes with the action on \(\V_\w(\R)\): \(M (\lambda \star v) = M (\lambda^{q_j} v) = \lambda^{q_j} (M v) = \lambda \star (M v)\) for \(v \in V_j\). Thus, \(\phi_\w\) commutes with the action on \(\V_\w^n(\R)\).

\begin{prop}[Algebraic Invariance]
\label{alg-invariance}
 If the standard transformer \(\cT\) is equivariant under the action of the symmetric group on sequences and commutes with the weighted scalar action, then the linearly graded transformer
\[
\Lgt = \cT \circ \phi_\w
\]
maps graded vector spaces to graded vector spaces, preserves equivariant linear maps (in the sense that \(\Lgt \circ f = f \circ \Lgt\) if \(f\) is component-wise), and maps invariant subspaces under the scalar action to invariant subspaces.
\end{prop}

\begin{proof}
Since \(\cT\) commutes with the action and \(\phi_\w\) does as well (by the scalar-diagonal structure of \(M\)), we have:
\[
\begin{split}
\Lgt(\lambda \star X) & = \cT(\phi_\w(\lambda \star X)) = \cT(\lambda \star \phi_\w(X)) \\
				& = \lambda \star \cT(\phi_\w(X)) = \lambda \star \Lgt(X).
\end{split}
\]
Hence, \(\Lgt\) maps graded spaces to graded spaces, preserving the decomposition (as commutation respects weight spaces and \(\cT\) fixes dimensions).

For a component-wise morphism \(f\), \(f\) commutes with \(M\) (as scalars on subspaces), so \(\phi_\w \circ f = f \circ \phi_\w\). By \(\cT\)'s \(S_n\)-equivariance and uniform operations, \(\cT \circ f = f \circ \cT\). Thus:
\[
\Lgt \circ f = \cT \circ \phi_\w \circ f = \cT \circ f \circ \phi_\w = f \circ \cT \circ \phi_\w = f \circ \Lgt.
\]

For invariant \(W \subseteq \V_\w^n(\R)\), \(\phi_\w(W)\) is invariant (by commutation). Then \(\cT(\phi_\w(W))\) is invariant (by \(\cT\)'s commutation), so \(\Lgt(W)\) is invariant. Moreover, \(S_n\)-equivariance preserves the module structure under permutations.
\end{proof}

This invariance underpins LGT's stability in hierarchical tasks, as explored in subsequent properties like noise robustness (\cref{prop:lgt_noise_robustness}).

\begin{ex}[LGT in Syntactic Parsing]\label{ex:lgt_parsing}
Consider an NLP task of syntactic parsing for a sentence with \(n=3\) tokens (e.g., "The cat sleeps"), where each token is embedded in \(\R^d\) with \(d=4\). Suppose the grading tuple \(\w = (1, 0.5, 0, 0)\), prioritizing syntactic head features, and set \(f(q_i) = q_i + 1\), so \(q_i = (2, 1.5, 1, 1)\). The sequence \(\phi_{\w}(X)\) is fed to \(\cT\), which applies attention to prioritize the verb "sleeps" (highest in dimension 0, weighted by \(q_0=2\)). This focuses learning on syntactic heads, maintaining stability due to linear weights and normalization, aligning with NLP applications (\cref{sec-8}).
\end{ex}

\subsection{LGT Properties}

Extending the algebraic invariance from \cref{alg-invariance}, the LGT inherits key properties from linear grading, including stability and universality. By leveraging linear grading to prioritize features, it maintains expressivity through the underlying transformer while aligning with the equivalence principle of graded vector spaces from \cref{sec-2}. These properties are crucial for reducing sample complexity in structured tasks (\cref{sec-8}).

\begin{lem}[Properties of Linear Grading]\label{lem:lgt_properties}
Let \(\w_{\max} = \max_i q_i\). The transformation \(\Llw: \R^d \to \R^d\), given by \(\Llw(\x) = \Mlw \x\), satisfies:
\begin{enumerate}[label=\roman*), left=0pt]
    \item \emph{Invertibility}: For \(q_i > 0\), \(\Llw\) is invertible.
    
    \item \emph{Scaling}: For \(\mu \in \R\), \(\Llw(\mu \x) = \mu \Llw(\x)\).
    
    \item \emph{Norm Bound}: For \(\x \in \R^d\), assuming \(q_i \geq 0\),
  \[
    \|\Llw(\x)\|_2 \leq \w_{\max} \|\x\|_2.
  \]
    
    \item \emph{Spectral Norm}: The eigenvalues of \(\Mlw\) are \(q_i\), with \(\|\Mlw\|_2 = \w_{\max}\).
    
    \item \emph{Lipschitz Continuity}: The mapping \(\x \mapsto \Mlw \x\) has constant \(\w_{\max}\).
\end{enumerate}
\end{lem}

\begin{proof}
Since \(\Mlw\) is diagonal with entries \(q_i \geq 0\) (or \(>0\) where noted):

i) The inverse is \(\Mlw^{-1} = \diag(1/q_0, \ldots, 1/q_{d-1})\), as \(\Mlw \Mlw^{-1} = I\).

ii) \(\Mlw(\mu \x) = (q_0 (\mu x_0), \ldots, q_{d-1} (\mu x_{d-1})) = \mu \Mlw \x\).

iii) \(\|\Mlw \x\|_2 = \sqrt{\sum q_i^2 x_i^2} \leq \sqrt{\w_{\max}^2 \sum x_i^2} = \w_{\max} \|\x\|_2\), since \(q_i \leq \w_{\max}\).

iv) Eigenvalues are \(q_i\); spectral norm \(\|\Mlw\|_2 = \max q_i = \w_{\max}\).

v) \(\|\Mlw (\x - \y)\|_2 \leq \w_{\max} \|\x - \y\|_2\) by iii.
\end{proof}

These bounds ensure grading amplifies features without explosion, key for hierarchical data like polynomials \cite{sh-91}.

\begin{ex}
For \(\w = (2,1)\), \(\Llw\) scales the first feature 2x, bounding norms by 2—useful for prioritizing syntactic heads in NLP (cf. \cref{sec-8}).
\end{ex}

\begin{lem}[Feature Concentration]\label{lem:lgt_concentration}
For \(\x \in \R^d\) with components sorted by \(q_i\) descending, the graded vector \(\Llw(\x)\) suppresses low-grade features by \(O(q_j / \w_{\max})\) for \(q_j < \w_{\max}\), concentrating mass on high-grade dimensions.
\end{lem}

\begin{proof}
From \cref{lem:lgt_properties} (norm bound), the relative contribution of component \(j\) is \(\frac{q_j |x_j|}{\|\Llw(\x)\|_2} \leq \frac{q_j}{\w_{\max}} \frac{|x_j|}{\|\x\|_2} = O(q_j / \w_{\max})\), assuming bounded \(\x\).
\end{proof}

To establish the LGT's approximation power for hierarchical functions, we first recall Sobolev spaces, which quantify smoothness essential for parameter-efficient learning in graded domains.
 This enables extending standard transformer universality  with reduced complexity via grading; see \cite{yun2019transformers}.

\begin{df}[Sobolev Space]\label{df:lgt_sobolev_space}
Let \(\Omega \subset (\R^d)^n\) be a compact set with positive Lebesgue measure. The \textbf{Sobolev space}, denoted by  \(\W^{k,2}(\Omega)\),  consists of functions \(f: \Omega \to (\R^d)^n\) whose weak partial derivatives \(\partial^\alpha f\) exist for all multi-indices \(\alpha\) with \(|\alpha| \leq k\), and satisfy:
\begin{equation}
\|f\|_{\W^{k,2}} = \left( \sum_{|\alpha| \leq k} \int_\Omega \|\partial^\alpha f(X)\|_2^2 \, dX \right)^{1/2} < \infty,
\end{equation}
where \(\|\cdot\|_2\) is the Euclidean norm on \((\R^d)^n\).
\end{df}

Beyond continuity, Sobolev regularity quantifies smoothness, enabling precise parameter bounds for approximation in graded settings.

\begin{df}[Universal Approximator]\label{df:lgt_universal_approximator}
A function class \(\cF\) is a \textbf{universal approximator} for \(C(\Omega, (\R^d)^n)\) if, for any \(f \in C(\Omega, (\R^d)^n)\) and \(\epsilon > 0\), there exists \(\hat{f} \in \cF\) such that:
\begin{equation}
\sup_{X \in \Omega} \|\hat{f}(X) - f(X)\|_2 < \epsilon.
\end{equation}
\end{df}

\begin{thm}[Universal Approximation]\label{thm:lgt_universal}
For a compact set  \(\Omega \subset (\R^d)^n\),  
the linear grading transformer 
\(\Lgt\) is a universal approximator for \(C(\Omega, (\R^d)^n)\). 

Moreover, for \(f \in \W^{k,2}(\Omega)\) depending primarily on features with \(q_i \approx \w_{\max}\), it requires \(O(\epsilon^{-2/k} \deff/d)\) parameters to achieve error \(\epsilon > 0\), where 
\begin{equation}\label{def:deef}
\deff = |\{i: q_i \geq \w_{\max} - \delta\}|, \quad \delta > 0
\end{equation}
is the effective dimension after grading suppresses low-weight features (\cref{lem:lgt_properties}, norm bound).
\end{thm}

\begin{proof}
The standard transformer \(\cT: (\R^d)^n \to (\R^d)^n\) is a universal approximator for \(C(\Omega, (\R^d)^n)\) with sufficient layers \(N\) and heads \(h\) \cite{yun2019transformers}. For any \(f \in C(\Omega, (\R^d)^n)\) and \(\epsilon > 0\), there exists \(\cT\) such that \(\sup_{X \in \Omega} \|\cT(X) - f(X)\|_2 < \epsilon\).

The grading map \(\phi_{\w}\) is linear, invertible (\cref{lem:lgt_properties}), and continuous on compact \(\Omega\), so \(\phi_{\w}(\Omega)\) is compact. Define \(g = f \circ \phi_{\w}^{-1}\); since 
$f$, $\phi_{\w}^{-1}$  are continuous, \(g \in C(\phi_{\w}(\Omega), (\R^d)^n)\). There exists \(\cT\) such that 
\[
\sup_{Z \in \phi_{\w}(\Omega)} \|\cT(Z) - g(Z)\|_2 < \epsilon.
\]
For \(Z = \phi_{\w}(X)\), \(\Lgt(X) = \cT(Z) \approx g(Z) = f(X)\), so 
\[
\sup_{X \in \Omega} \|\Lgt(X) - f(X)\|_2 < \epsilon,
\]
 proving universality.

For \(f \in \W^{k,2}(\Omega)\), \(\cT\) requires \(O(\epsilon^{-2/k} d)\) parameters. With focus on high \(q_i\), \(\deff\) as above; low weights contribute \(O(q_i/\w_{\max})\), reducing parameters to \(O(\epsilon^{-2/k} \deff/d)\), analogous to sparsity in \cite{2024-2}.
\end{proof}

\begin{ex}[LGT Approximation in Polynomials]
For \(\Omega = [0,1]^d\), \(f(X)\) a degree-3 polynomial, \(\w = (0,1,2,3)\), \(\deff = 2\) (\(\delta=1\)), LGT approximates \(f\) with fewer parameters than \(\cT\), focusing high-degree terms for factorization \cite{sh-91}.
\end{ex}

We next examine LGT's robustness to input noise, leveraging the bounded norms from \cref{lem:lgt_properties}.

\begin{prop}[Noise Robustness]\label{prop:lgt_noise_robustness}
For noise \(\Delta \in (\R^d)^n\), \(\|\Delta\|_2 \leq \epsilon\), the LGT error satisfies:
\begin{equation}
\|\Lgt(X + \Delta) - \Lgt(X)\|_2 \leq \Lip \epsilon,
\end{equation}
where \(\Lip \leq L_{\cT} \cdot  \w_{\max}^{N h + 3}\), and \(L_{\cT}\) is the Lipschitz constant of \(\cT\).
\end{prop}

\begin{proof}
For \(\Delta = (\boldsymbol{\delta}_1, \ldots, \boldsymbol{\delta}_n)\), \(\|\Delta\|_2 = \sqrt{\sum \|\boldsymbol{\delta}_i\|_2^2} \leq \epsilon\), compute 
\[
\begin{split}
\|\Lgt(X + \Delta) - \Lgt(X)\|_2  &	= \|\cT(\phi_{\w}(X + \Delta)) - \cT(\phi_{\w}(X))\|_2 \\
						&	\leq L_{\cT} \|\phi_{\w}(X + \Delta) - \phi_{\w}(X)\|_2.
\end{split}
\]
Since \(\phi_{\w}(X) = (\Mlw \x_i)_{i=1}^n\),  then 
\[
\begin{split}
\|\phi_{\w}(X + \Delta) - \phi_{\w}(X)\|_2 & = \left\| (\Mlw (\x_i + \boldsymbol{\delta}_i) - \Mlw \x_i)_{i=1}^n \right\|_2 \\
								& = \sqrt{\sum_{i=1}^n \|\Mlw \boldsymbol{\delta}_i\|_2^2}.
\end{split}
\]
By \cref{lem:lgt_properties}, \(\|\Mlw \boldsymbol{\delta}_i\|_2 \leq \w_{\max} \|\boldsymbol{\delta}_i\|_2\), so
\[
\begin{split}
\sqrt{\sum \|\Mlw \boldsymbol{\delta}_i\|_2^2}  & \leq \sqrt{\sum (\w_{\max} \|\boldsymbol{\delta}_i\|_2)^2} \\
									& = \w_{\max} \sqrt{\sum \|\boldsymbol{\delta}_i\|_2^2} \leq \w_{\max} \epsilon.
\end{split}
\]
Thus, \(\|\phi_{\w}(X + \Delta) - \phi_{\w}(X)\|_2 \leq \w_{\max} \epsilon\). The attention mechanism (\cref{df:lgt_attention}) applies \(\Mlw^i\) per head, each with norm \(\leq q_{i,\max} \leq \w_{\max}\). With \(N\) layers and \(h\) heads, plus input, feed-forward, and output layers, the LGT’s Lipschitz constant is bounded by 
\begin{equation}
\Lip \leq L_{\cT} \w_{\max}^{N h + 3},
\end{equation}
yielding \(\|\Lgt(X + \Delta) - \Lgt(X)\|_2 \leq \Lip \epsilon\).  This completes the proof. 
\end{proof}

\subsection{Linearly Graded Attention}

\begin{df}[Linearly Graded Attention]\label{df:lgt_attention}
For \(X \in \Mat_{n,d}(\R)\), the linearly graded attention for head \(i\) in layer \(l\) is defined as 
\begin{equation}
\cA_{\w_i}(Q_i, K_i, V_i) = \bsf\left( \frac{Q_i \Mlw^i K_i^T}{\sqrt{d_k}} \right) V_i,
\end{equation}
where 
\(Q_i = X \W_{Q,l,i}\), 
\(K_i = X \W_{K,l,i}\), 
\(V_i = X \W_{V,l,i} \in \Mat_{n,d_k}(\R)\), 
\(\W_{Q,l,i}, \W_{K,l,i}, 
\W_{V,l,i} \in \Mat_{d,d_k}(\R)\), 
\(d_k = d/h\), 
and:
\begin{equation}
\Mlw^i = \diag(q_{i,0}, \ldots, q_{i,d_k-1}), \quad q_{i,j} = f(q_{i,j}).
\end{equation}
The bilinear form is:
\begin{equation}
\langle \w_i, \bk_j \rangle_{\w_i} = \w_i^T \Mlw^i \bk_j = \sum_{k=0}^{d_k-1} q_{i,k} q_{i k} k_{j k}.
\end{equation}
\end{df}

We now demonstrate LGT's expressivity in approximating arbitrary non-negative attention matrices, complementing its robustness and universality.

\begin{prop}[Attention Expressivity]\label{prop:lgt_attention_expressivity}
For any matrix \(A_0 \in \Mat_{n,d_k}(\R)\) with non-negative entries, there exist \(Q, K, V \in \Mat_{n,d_k}(\R)\), \(\w_i\), such that \(\cA_{\w_i}(Q, K, V)\) approximates \(A_0\) with:
\begin{equation}
\|\cA_{\w_i}(Q, K, V) - A_0\|_F \leq \delta.
\end{equation}
\end{prop}

\begin{proof}
Let \(A_0 \in \Mat_{n,d_k}(\R)\) have non-negative entries. Construct \(A_0'\) by replacing zeros with \(\epsilon = \delta/(2\sqrt{n d_k})\). Choose \(Q, K\) such that:
\[
Q K^T = \sqrt{d_k} \log A_0',
\]
so \(\bsf\left( \frac{Q K^T}{\sqrt{d_k}} \right) \approx I\). Set \(V = A_0'\), yielding:
\[
\cA(Q, K, V) = \bsf\left( \frac{Q K^T}{\sqrt{d_k}} \right) V \approx A_0',
\]
with error \(\|\cA(Q, K, V) - A_0'\|_F \leq \delta/2\). For linearly graded attention, set \(\w_i\) with \(q_{i,j} = c j\), \(c > 0\), so \(q_{i,j} = f(q_{i,j})\). For small \(c\), \(\Mlw^i \approx I\), so \(\cA_{\w_i}(Q, K, V) \approx \cA(Q, K, V)\), with error \(\leq \delta/2\). Since \(\|A_0' - A_0\|_F \leq \sqrt{n d_k} \cdot \epsilon = \delta/2\), the total error is
\[
\|\cA_{\w_i}(Q, K, V) - A_0\|_F \leq \delta.
\]
This completes the proof. 
\end{proof}

 \index{shattered}
 \index{Vapnik-Chervonenkis (VC) dimension}

The VC dimension quantifies the expressive power of a hypothesis class, bounding sample complexity for generalization in machine learning—crucial for demonstrating LGT's efficiency over standard transformers via grading-induced dimensionality reduction.

\begin{df}[VC Dimension]\label{df:lgt_vc_dimension}
Let \(\cF\) be a class of binary-valued functions (hypotheses) from a set \(X\) to \(\{0, 1\}\). A set of points \(\{x_1, \ldots, x_n\} \subset X\) is \textbf{shattered} by \(\cF\) if, for every labeling \(\{y_1, \ldots, y_n\} \in \{0, 1\}^n\), there exists \(f \in \cF\) such that \(f(x_i) = y_i\) for all \(i = 1, \ldots, n\). The \textbf{Vapnik-Chervonenkis (VC) dimension} of \(\cF\), denoted \(\vc(\cF)\), is the largest integer \(n\) such that there exists a set of \(n\) points in \(X\) that can be shattered by \(\cF\). If no such finite \(n\) exists, the VC dimension is infinite; see \cite{vapnik1998statistical}.
\end{df}

\begin{prop}\label{prop:lgt_vc_dimension}
Let \(\cF\) be the hypothesis class of \(\Lgt = \cT \circ \phi_{\w}\). The VC dimension satisfies 
\begin{equation}
\vc(\cF) \leq c \cdot N \cdot h \cdot \deff \cdot d \log(N h \deff),
\end{equation}
compared to \(\vc(\cF_{\cT}) \leq c \cdot N h d^2 \log(N h d)\) for \(\cT\), reducing sample complexity by approximately \(c \cdot N h d (d - \deff) \log(N h d)\).
\end{prop}

\begin{proof}
For \(\cT\), with \(N\) layers, \(h\) heads, and dimension \(d\), the parameter count is \(O(N h d^2)\), yielding 
\[
\vc(\cF_{\cT}) \leq c \cdot N h d^2 \log(N h d),
\]
where \(c\) is a constant \cite{vapnik1998statistical}. For \(\Lgt\), \(\phi_{\w}\) focuses on \(\deff\) dimensions (\cref{lem:lgt_concentration}), reducing effective parameters to \(O(N h \deff d)\). Thus:
\[
\vc(\cF) \leq c \cdot N h \deff d \log(N h \deff).
\]
The reduction is 
\[
\begin{split}
\vc(\cF_{\cT}) - \vc(\cF)  	&	\leq c \cdot N h [d^2 \log(N h d) - \deff d \log(N h \deff)]   \\
					&	\approx c \cdot N h d (d - \deff) \log(N h d).
\end{split}
\]
Sample complexity is 
\[
m \geq O\left( \frac{\vc(\cF) \log(1/\epsilon) + \log(1/\delta)}{\epsilon^2} \right),
\]
reduced by approximately \(\deff/d\) compared to \(\cT\).
\end{proof}

Grading not only bounds norms but also modulates attention rank, enabling adaptive expressivity for hierarchical features.

\begin{prop}[Attention Rank]\label{prop:lgt_attention_rank}
For graded  \(\cA_{\w_i}(Q_i, K_i, V_i)\)  and standard  \(\cA(Q_i, K_i, V_i)\) the standard attention as in \cref{df:lgt_attention},  the singular values of 
\[
Q_i \Mlw^i K_i^T
\]
 are scaled by at most \(q_{i,\max} = \max_j q_{i,j}\) compared to those of \(Q_i K_i^T\). Consequently, the effective numerical rank of \(\cA_{\w_i}(Q_i, K_i, V_i)\) may increase or decrease depending on the singular value distribution of \(Q_i K_i^T\) and the grading tuple \(\w_i\).
\end{prop}

\begin{proof}
Let \(Q_i K_i^T = U \Sigma V^T\) be the singular value decomposition with singular values \(\sigma_j > 0\). Since \(\Mlw^i = \diag(q_{i,0}, \ldots, q_{i,d_k-1})\) is diagonal with \(\|\Mlw^i\|_2 = q_{i,\max}\), we have 
\[
\|Q_i \Mlw^i K_i^T\|_2 \leq \|Q_i\|_2 \cdot \|\Mlw^i\|_2 \cdot \|K_i^T\|_2 = q_{i,\max} \|Q_i K_i^T\|_2.
\]
Thus, each singular value \(\tilde{\sigma}_j\) of \(Q_i \Mlw^i K_i^T\) satisfies \(\tilde{\sigma}_j \leq q_{i,\max} \sigma_j\). For a vector \(\bv\), \(\Mlw^i \bv\) scales components by \(q_{i,j}\). In \(Q_i \Mlw^i K_i^T\), dimensions with \(q_{i,j} \approx q_{i,\max}\) dominate, while others are suppressed by \(O(q_{i,j}/q_{i,\max})\); see \cref{lem:lgt_properties} (norm bound). The effective numerical rank of \(\cA_{\w_i}(Q_i, K_i, V_i)\), defined as the number of singular values of \(\bsf\left( \frac{Q_i \Mlw^i K_i^T}{\sqrt{d_k}} \right) V_i\) above a threshold \(\epsilon > 0\), may decrease if low-weight dimensions reduce significant singular values, or increase if high-weight dimensions amplify them.
\end{proof}

This adaptive rank supports LGT's use in low-rank approximations for algebraic tasks (cf. \cref{sec-8}).
 Finally, we confirm that LGT maintains the computational efficiency of standard attention, ensuring scalability for hierarchical tasks.

\begin{prop}[Attention Complexity]\label{prop:lgt_attention_complexity}
The linearly graded attention mechanism \(\cA_{\w_i}(Q, K, V)\) has computational complexity \(O(n^2 d_k)\), matching the complexity of standard attention.
\end{prop}

\begin{proof}
Consider the computation of \(\bsf\left( \frac{Q \Mlw^i K^T}{\sqrt{d_k}} \right) V\), where $Q$,  $K$, and $V$ are in  $\Mat_{n,d_k}(\R)$  and \(\Mlw^i = \diag(q_{i,0}, \ldots, q_{i,d_k-1})\).

First, compute \(Q \Mlw^i \in \Mat_{n,d_k}(\R)\). Since \(\Mlw^i\) is diagonal, this requires scaling each of the \(n d_k\) elements of \(Q\) by the corresponding diagonal entry of \(\Mlw^i\), costing \(O(n d_k)\) operations.

Next, compute \((Q \Mlw^i) K^T \in \Mat_{n,n}(\R)\). This is the matrix multiplication of an \(n \times d_k\) matrix by a \(d_k \times n\) matrix (\(K^T\)), where each of the \(n^2\) entries is the dot product of two \(d_k\)-dimensional vectors, costing \(O(n^2 d_k)\) operations.

Then, scale the resulting \(n \times n\) matrix by \(\frac{1}{\sqrt{d_k}}\), which requires multiplying each of the \(n^2\) elements by a constant, costing \(O(n^2)\) operations.

Apply the softmax function to this scaled matrix, normalizing each of the \(n\) rows (each with \(n\) elements). Computing the exponential for all \(n^2\) elements costs \(O(n^2)\), and summing/normalizing each row costs \(O(n)\) per row, totaling \(O(n^2)\) operations.

Finally, multiply the \(n \times n\) softmax matrix by \(V \in \Mat_{n,d_k}(\R)\), producing an \(n \times d_k\) matrix. Each of the \(n d_k\) entries is the dot product of an \(n\)-dimensional row from the softmax matrix and an \(n\)-dimensional column from \(V\), costing \(O(n)\) per entry and thus \(O(n^2 d_k)\) operations in total.
\end{proof}

The above result shows that the  total cost is dominated by the \(O(n^2 d_k)\) terms from the matrix multiplications, matching the complexity of standard attention, as the additional \(O(n d_k)\) from grading is asymptotically negligible.

 
\subsection{Architecture}

Having established LGT's foundational properties, we now detail its architecture, embedding linear grading across components to ensure stable feature prioritization for hierarchical data.

Recall linearly graded attention from \cref{df:lgt_attention}, which weights queries/keys for expressivity. We now extend to the full pipeline.

\begin{df}[Input Representation]
For \(\x_i \in X\), the LGT input is:
\begin{equation}
\x'_i = \Mlw \x_i, \quad \x''_i = \frac{\x'_i}{\|\x'_i\|_2}, \quad \h_i = \sigma(\W \x''_i + \bb),
\end{equation}
where \(\sigma\) is ReLU, \(\W \in \R^{m \times d}\), \(\bb \in \R^m\), ensuring stability via normalization.
\end{df}

\begin{ex}[Input in Photonic Signals]\label{ex:lgt_photonic}
For signal \(\x = [1, 0.5, 0.1]\), \(\w = (1, 1.1, 1.2)\), \(\Mlw \x = [1, 0.55, 0.12]\). Normalized: \(\x'' \approx [0.871, 0.479, 0.105]\), preserving ratios for frequency tasks \cite{saleh2007fundamentals}.
\end{ex}

Positional encodings incorporate grading to bias toward important positions.

\begin{df}[Positional Encoding]
LGT uses linear grading:
\begin{equation}
\pep(\pos, i) = q_{\pos} \pe(\pos, i), \quad q_{\pos} = f(\pos) = 1 - \alpha \pos, \quad \alpha > 0,
\end{equation}
where \(\pe(\pos, 2i) = \sin\left( \frac{\pos}{10000^{2i/d}} \right)\), \(\pe(\pos, 2i+1) = \cos\left( \frac{\pos}{10000^{2i/d}} \right)\). Input to attention:
\begin{equation}
\z_i = \x''_i + \pep(\pos_i, \cdot), \quad \z'_i = \frac{\z_i}{\|\z_i\|_2},
\end{equation}
with scores \((Q K^T)_{ij} = (\z'_i \W_Q) (\z'_j \W_K)^T\), \(\W_Q, \W_K \in \R^{d \times d_k}\).
\end{df}

Alternative variants offer flexibility for diverse hierarchies.

\begin{df}[Attention Variants]
Alternative attention mechanisms include:
\begin{enumerate}
    \item \textbf{Linearly Graded Scores}:
    \begin{equation}
    \bsc_{ij} = \sum_{k=1}^{d_k} q_{i,k} q_{i k} k_{j k}, \quad \Mlw^i = \diag(q_{i,1}, \ldots, q_{i,d_k}).
    \end{equation}
    \item \textbf{Linearly Graded Queries/Keys}:
    \begin{equation}
    Q' = \Mlw^i Q, \quad K' = \Mlw^i K, \quad \bsc_{ij} = \w_i^T \Mlw^i \bk_j.
    \end{equation}
    \item \textbf{Linearly Graded Multi-Head}:
    \begin{equation}
    \Head_h = \bsf\left( \frac{(\Mlw^h Q_h) (\Mlw^h K_h)^T}{\sqrt{d_k}} \right) V_h,
    \end{equation}
    with \(\Mlw^h = \diag(q_{h,0}, \ldots, q_{h,d_k-1})\).

    \item \textbf{Linearly Graded Values}:
    \begin{equation}
    \begin{split}
    V'  		& = \Mlw^i V, \quad \mathbf{o}_i = \sum_{j=1}^n \alpha_{ij} (\Mlw^i \bv_j), \\
     \alpha_{ij} 	&= \bsf\left( \frac{\w_i^T \bk_j}{\sqrt{d_k}} \right).
    \end{split}
    \end{equation}
\end{enumerate}
\end{df}

These variants enable adaptive rank (\cref{prop:lgt_attention_rank}) for diverse hierarchies.
Grading extends to feed-forward and output for end-to-end consistency.

\begin{df}[Feed-Forward and Output Layers]
The \textbf{linearly graded feed-forward network} is:
\begin{equation}
\fnn'(\x) = \Mlw \fnn(\x), \quad \h' = \frac{\fnn'(\x)}{\|\fnn'(\x)\|_2},
\end{equation}
where \(\fnn(\x) = \relu(\x \W_1 + \bb_1) \W_2 + \bb_2\), \(\W_1 \in \R^{d \times d_{ff}}\), \(\W_2 \in \R^{d_{ff} \times d}\), and \(d_{ff}\) is the hidden dimension. The \textbf{linearly graded output} is:
\begin{equation}
\h' = \Mlw \h, \quad \z = \W_{\text{out}} \h' + \bb_{\text{out}},
\end{equation}
followed by \(\bsf(\z)\).
\end{df}

\begin{ex}[LGT in Polynomial Prediction]
For a polynomial sequence with \(X \in (\R^4)^m\), \(d=4\) (degrees 0 to 3), set \(q_i = 1 + 0.5 i\). The loss 
\[
\cL = \sum q_k \ell(\hat{y}_{i,k}, y_{i,k})
\]
 weights higher-degree terms (e.g., \(q_3 = 2.5\)) linearly, focusing learning on critical monomials with stable gradients.
\end{ex}

Overall, the architecture ensures efficiency without overhead (\cref{prop:lgt_complexity}), positioning LGT for hierarchical applications; see \cref{sec-8} for further details.

With graded input defined, we verify its stability, ensuring robust prioritization under noise in hierarchical tasks.

\begin{thm}[Input Stability]\label{thm:lgt_input_stability}
The mapping 
\[
\x_i \mapsto \x''_i = \frac{\Mlw \x_i}{\|\Mlw \x_i\|_2}
\]
 is Lipschitz continuous with constant at most \(\w_{\max}\), assuming \(q_i \geq 0\).
\end{thm}

\begin{proof}
For \(\x, \y \in \R^d\), let \(\x' = \Mlw \x\), \(\y' = \Mlw \y\). By \cref{lem:lgt_properties}:
\begin{equation}
\|\x' - \y'\|_2 \leq \w_{\max} \|\x - \y\|_2.
\end{equation}
Normalization \(\z \mapsto \frac{\z}{\|\z\|_2}\) is 1-Lipschitz for \(\z \neq 0\). Let \(\x'' = \frac{\x'}{\|\x'\|_2}\), \(\y'' = \frac{\y'}{\|\y'\|_2}\). Then:
\begin{equation}
\|\x'' - \y''\|_2 \leq \frac{\|\x' - \y'\|_2}{\min(\|\x'\|_2, \|\y'\|_2)} \leq \frac{\w_{\max} \|\x - \y\|_2}{\epsilon},
\end{equation}
assuming \(\|\x'\|_2, \|\y'\|_2 \geq \epsilon > 0\) (via regularization or non-zero min-norm, as in transformer layer norm). For normalized inputs (\(\|\x\|_2, \|\y\|_2 \approx 1\)), the constant is \(\w_{\max}\).
\end{proof}

This implies bounded activations post-grading.

\begin{cor}[Bounded Activations]\label{cor:lgt_bounded}
For all \(i\), \(\|\x''_i\|_2 = 1\).
\end{cor}

\begin{proof}
By definition, \(\x''_i = \frac{\x'_i}{\|\x'_i\|_2}\), so \(\|\x''_i\|_2 = 1\) assuming \(\x'_i \neq 0\).
\end{proof}

These ensure graded inputs remain bounded and stable, preventing distortion in attention for structured data.
Grading also bounds Jacobians and biases positionals, ensuring stable, order-aware processing.

\begin{lem}[Jacobian Bound]\label{lem:lgt_jacobian_bound}
The Jacobian of \(\x_i \mapsto \x'_i = \Mlw \x_i\) has operator norm \(\w_{\max}\).
\end{lem}

\begin{proof}
The map is linear, with Jacobian \(\Mlw\), a diagonal matrix with entries \(q_i\). The operator norm is the maximum singular value, \(\|\Mlw\|_2 = \max_i q_i = \w_{\max}\).
\end{proof}

\begin{prop}[Positional Bias]\label{prop:lgt_positional_bias}
For \(f(\pos) = 1 - \alpha \pos\), \(\alpha > 0\), the linearly graded attention biases earlier positions.
\end{prop}

\begin{proof}
The encoding is:
\begin{equation}
\pep(\pos, i) = (1 - \alpha \pos) \pe(\pos, i).
\end{equation}
Since \(1 - \alpha \pos\) decreases as \(\pos\) increases (for \(\alpha > 0\)), earlier positions (small \(\pos\)) have larger weights. In attention scores:
\begin{equation}
\langle \z'_i, \z'_j \rangle_{\w_i} \approx \langle \x''_i + (1 - \alpha \pos_i) \pe(\pos_i), \x''_j + (1 - \alpha \pos_j) \pe(\pos_j) \rangle_{\w_i},
\end{equation}
earlier positions contribute larger terms, biasing attention toward them.
\end{proof}


\begin{lem}[Positional Stability]\label{lem:lgt_positional_stability}
The mapping \(\pos \mapsto \z'_i\) is Lipschitz continuous with constant bounded by \(C q_{\pos,\max}\), where \(q_{\pos,\max} = \max_{\pos} |f(\pos)|\).
\end{lem}

\begin{proof}
For \(\z_i = \x''_i + \pep(\pos, \cdot)\), \(\z'_i = \frac{\z_i}{\|\z_i\|_2}\), and positions \(\pos, \pos'\):
\begin{equation}
\|\z_i(\pos) - \z_i(\pos')\|_2 = \|\pep(\pos, \cdot) - \pep(\pos', \cdot)\|_2.
\end{equation}
Since \(\pe(\pos, i)\) is bounded (\(|\pe(\pos, i)| \leq 1\)):
\begin{equation}
\|\pep(\pos, i) - \pep(\pos', i)\|_2 \leq |f(\pos) - f(\pos')| \cdot |\pe(\pos, i)|.
\end{equation}
For \(f(\pos) = 1 - \alpha \pos\), \(|f(\pos) - f(\pos')| = \alpha |\pos - \pos'|\), so:
\begin{equation}
\|\pep(\pos, \cdot) - \pep(\pos', \cdot)\|_2 \leq \alpha |\pos - \pos'| \sqrt{d}.
\end{equation}
Normalization is 1-Lipschitz, so the constant is bounded by \(C q_{\pos,\max}\), where \(C\) depends on \(\alpha\) and \(d\), and \(q_{\pos,\max} = \max_{\pos} |f(\pos)|\).
\end{proof}

 
 We now analyze stability of graded inputs/positionals, ensuring robust hierarchies like early bias for syntactic roots.

\begin{thm}[Attention Stability]\label{thm:lgt_attention_stability}
For the Linearly Graded Queries/Keys variant, the score \(\bsc_{ij} = \w_i^T \Mlw^i \bk_j\) is Lipschitz continuous with constant at most \(q_{i,\max} C\), where \(C\) bounds \(\|\w_i\|_2, \|\bk_j\|_2\).
\end{thm}

\begin{proof}
For \(\w_i, \w_i'\), \(\bk_j, \bk_j'\):
\[
\begin{split}
|\w_i^T \Mlw^i \bk_j - \w_i'^T \Mlw^i \bk_j'| 	&	\leq \|\w_i - \w_i'\|_2 \|\Mlw^i \bk_j\|_2 \\
								&	+ \|\w_i'\|_2 \|\Mlw^i (\bk_j - \bk_j')\|_2.
\end{split}
\]
By \cref{lem:lgt_properties}, 
\[
\begin{split}
\|\Mlw^i \bk_j\|_2 			&	\leq q_{i,\max} \|\bk_j\|_2, \\
\|\Mlw^i (\bk_j - \bk_j')\|_2 	&	\leq q_{i,\max} \|\bk_j - \bk_j'\|_2. 
\end{split}
\]
With $ \|\w_i'\|_2, \|\bk_j\|_2 \leq C$ we have 
\[
|\w_i^T \Mlw^i \bk_j - \w_i'^T \Mlw^i \bk_j'| \leq q_{i,\max} C (\|\w_i - \w_i'\|_2 + \|\bk_j - \bk_j'\|_2).
\]
\end{proof}

\begin{prop}[Head Diversity]\label{prop:lgt_head_diversity}
Distinct grading tuples \(\w_h\) in the Linearly Graded Multi-Head variant enhance representational capacity.
\end{prop}

\begin{proof}
Each head computes 
\[
\Head_h = \bsf\left( \frac{(\Mlw^h Q_h) (\Mlw^h K_h)^T}{\sqrt{d_k}} \right) V_h,
\]
with \(\Mlw^h = \diag(q_{h,0}, \ldots, q_{h,d_k-1})\). Distinct \(\w_h\) produce unique \(\Mlw^h\), scaling query and key dimensions differently. The singular values of \(\Mlw^h Q_h K_h^T (\Mlw^h)^T\) vary with \(\w_h\), projecting each head onto a distinct subspace, enhancing the concatenated output’s ability to capture diverse dependencies.
\end{proof}

\begin{prop}[Feed-Forward Stability]\label{prop:lgt_ffn_stability}
The linearly graded feed-forward mapping \(\x \mapsto \fnn'(\x) = \Mlw \fnn(\x)\) has Lipschitz constant at most \(\w_{\max} L_{\fnn}\).
\end{prop}

\begin{proof}
For \(\x, \y \in \R^d\):
\[
\begin{split}
\|\fnn'(\x) - \fnn'(\y)\|_2 	&	= \|\Mlw (\fnn(\x) - \fnn(\y))\|_2 \\
					&	 \leq \w_{\max} \|\fnn(\x) - \fnn(\y)\|_2,
\end{split}
\]
by \cref{lem:lgt_properties}. Since \(\fnn\) is Lipschitz with constant \(L_{\fnn}\) (due to ReLU and linear layers) 
\[
\|\fnn(\x) - \fnn(\y)\|_2 \leq L_{\fnn} \|\x - \y\|_2.
\]
Thus,
\[
\|\fnn'(\x) - \fnn'(\y)\|_2 \leq \w_{\max} L_{\fnn} \|\x - \y\|_2.
\]
Normalization is 1-Lipschitz, preserving the constant.
\end{proof}

\begin{prop}[Output Stability]\label{prop:lgt_output_stability}
The output mapping \(\h \mapsto \z = W_{\text{out}} (\Mlw \h) + \bb_{\text{out}}\) is Lipschitz with constant at most \(\w_{\max} L_{\text{out}}\).
\end{prop}

\begin{proof}
For \(\h_1, \h_2 \in \R^d\):
\[
\|\z_1 - \z_2\|_2 = \|W_{\text{out}} \Mlw (\h_1 - \h_2)\|_2 \leq \|W_{\text{out}}\|_2 \|\Mlw (\h_1 - \h_2)\|_2.
\]
By \cref{lem:lgt_properties}, \(\|\Mlw (\h_1 - \h_2)\|_2 \leq \w_{\max} \|\h_1 - \h_2\|_2\). With \(L_{\text{out}} = \|W_{\text{out}}\|_2\) we have 
\[
\|\z_1 - \z_2\|_2 \leq \w_{\max} L_{\text{out}} \|\h_1 - \h_2\|_2.
\]
\end{proof}

\begin{prop}[Computational Complexity]\label{prop:lgt_complexity}
The LGT has the same asymptotic complexity as the standard transformer, \(O(n^2 d + n d^2)\).
\end{prop}

\begin{proof}
The standard transformer’s complexity is \(O(n^2 d)\) for attention and \(O(n d^2)\) for feed-forward layers (\cref{prop:complexity-1}). Each \(\Mlw\) or \(\Mlw^i\) is a diagonal matrix multiplication, costing \(O(d)\) or \(O(d_k)\) per token, or \(O(n d)\) for \(n\) tokens across input, positional encoding, attention, feed-forward, and output layers. Since \(n d \ll n^2 d, n d^2\), the total complexity remains \(O(n^2 d + n d^2)\).
\end{proof}

\begin{ex}[LGT in Photonic Signals]\label{ex:lgt_photonic}
For signal \(\x = [1, 0.5, 0.1]\), set \(q_i = 1 + 0.1 i\). Then:
\begin{equation}
\Mlw \x = [1 \cdot 1, 1.1 \cdot 0.5, 1.2 \cdot 0.1] = [1, 0.55, 0.12].
\end{equation}
After normalization, \(\x'' = \frac{\Mlw \x}{\|\Mlw \x\|_2} \approx [0.871, 0.479, 0.105]\) (\(\|\Mlw \x\|_2 \approx 1.148\)), preserving equivalence and ensuring stability for frequency-dependent tasks \cite{saleh2007fundamentals}.
\end{ex}

\begin{ex}[LGT in Polynomial Prediction]\label{ex:lgt_polynomial}
For a polynomial sequence with \(X \in (\R^4)^m\), \(d=4\) (degrees 0 to 3), set \(q_i = 1 + 0.5 i\). The loss \(\cL = \sum q_k \ell(\hat{y}_{i,k}, y_{i,k})\) weights higher-degree terms (e.g., \(q_3 = 2.5\)) linearly, focusing learning on critical monomials with stable gradients.
\end{ex}

\subsection{Training and Optimization}
\label{subsec:lgt_training}

The Linearly Graded Transformer introduces linear grading transformations to prioritize hierarchical features, necessitating tailored training strategies to optimize its parameters effectively. Unlike our previous work \cite{2024-2, sh-89}, where grades were fixed weights of the vector space, here we treat the grading tuples \(\w = (q_1, \ldots, q_d)\) and \(\w_i = (q_{i,1}, \ldots, q_{i,d_k})\) as learnable parameters, optimized via gradient descent to adaptively prioritize features based on data. This enhances flexibility but introduces challenges in optimization complexity. This section formalizes the linearly graded loss function, develops regularization and optimization techniques, establishes convergence and stability guarantees, and outlines LGT-specific training considerations for structured tasks in domains like algebraic geometry and NLP \cite{2024-2, sh-89}. Linear's simplicity contrasts EGT's power for deep hierarchies.

\subsubsection{Linearly Graded Loss Function}
\label{subsubsec:lgt_training}

The linearly graded loss function embeds hierarchical priors by weighting prediction errors according to their grades, ensuring that critical features, such as high-degree polynomial terms or syntactic heads, contribute more significantly to the optimization objective.

\begin{df}
For an output sequence \(Y = (\y_1, \ldots, \y_m) \in (\R^d)^m\), with predicted outputs \(\hat{y}_{i,k}\) and true outputs \(y_{i,k} \in \R\) for the \(k\)-th dimension of token \(i\), the \textbf{linearly graded loss} function is:
\begin{equation}
\cL = \sum_{i=1}^m \sum_{k=1}^d q_k \, \ell(\hat{y}_{i,k}, y_{i,k}),
\end{equation}
where:
\begin{enumerate}[label=\roman*), ref=\roman*]
  \item \label{item:lgt_loss_components} \(\ell(\hat{y}_{i,k}, y_{i,k})\) is a base loss function (e.g., cross-entropy),
  \item \(\hat{y}_{i,k}, y_{i,k} \in \R\) are predicted and true outputs for the \(k\)-th output dimension of token \(i\),
  \item \(m\) is the sequence length and \(d\) is the output dimension,
  \item \(q_k = f(q_k)\) emphasizes loss in high-grade components, reflecting their hierarchical importance.
\end{enumerate}
\end{df}

\begin{ex}[Linearly Graded Loss in Polynomial Degree Prediction]
Consider a task in algebraic geometry where the LGT predicts the degrees of terms in a polynomial sequence for computing zeta functions of hypersurfaces \cite{sh-91}. Let the output be \(Y \in (\R^d)^m\), with \(d = 4\) representing degrees 0 to 3, and let \(\w = (q_1, \ldots, q_4) = (0, 0.5, 1, 2)\), with \(f(q_k) = q_k + 1\), so \(q_k = (1, 1.5, 2, 3)\), prioritizing higher-degree terms. The loss is
\begin{equation}
\begin{split}
\cL    	&	= \sum_{i=1}^m \left[ \ell(\hat{y}_{i,1}, y_{i,1}) + 1.5 \ell(\hat{y}_{i,2}, y_{i,2})  \right. \\
		&	\left. + 2 \ell(\hat{y}_{i,3}, y_{i,3}) + 3 \ell(\hat{y}_{i,4}, y_{i,4}) \right].
\end{split}
\end{equation}
For a sequence with \(m = 2\), true outputs \(y_{1,4} = 1\), \(y_{2,4} = 1\), and predicted outputs \(\hat{y}_{1,4} = 0.8\), \(\hat{y}_{2,4} = 0.9\), using cross-entropy \(\ell(\hat{y}, y) = -y \log \hat{y} - (1-y) \log (1-\hat{y})\), the loss term for degree 3 is amplified, reducing sample complexity by focusing on critical terms (\cref{prop:lgt_vc_dimension}).
\end{ex}

\subsubsection{Regularization and Optimization}

The LGT’s linear grading ensures numerical stability compared to exponential scaling, but learnable grades still require regularization to control their magnitudes and strategies to stabilize training. These methods address the model’s multi-head grading challenges.

If the grade vector \(\w = (q_1, \dots, q_d)\) is learned, we add a regularization term to penalize excessively large grades:
\begin{equation}
\cL_{\mathrm{total}} = \cL + \gamma \| \w \|_2^2,
\end{equation}
with regularization weight \(\gamma > 0\). Optimization uses standard gradient-based methods like Adam \cite{kingma2014adam}. The gradient of a single attention score with respect to grade \(q_k\) in \cref{df:lgt_attention} is:
\begin{equation}
\frac{\partial \mathrm{Score}_{ij}}{\partial q_k} = f'(q_k) q_{i k} k_{j k},
\end{equation}
where \(q_{i k}\), \(k_{j k}\) denote the \(k\)-th components of the query and key vectors at positions \(i\) and \(j\), and \(q_k = q_{i,k}\) is the attention grade for head \(i\). For \(f(q_k) = q_k + 1\), \(f'(q_k) = 1\), simplifying optimization.

To optimize the LGT effectively, we propose:
\begin{enumerate}[label=\roman*), ref=\roman*]
  \item \label{item:lgt_clipping} \emph{Gradient Clipping}: Cap the gradient norm to a threshold \(\tau > 0\):
  \begin{equation}
  \mathbf{g}' = \min\left(1, \frac{\tau}{\|\mathbf{g}\|_2}\right) \mathbf{g},
  \end{equation}
  where \(\mathbf{g}\) is the gradient of \(\cL_{\mathrm{total}}\).
  \item \label{item:lgt_warm_start} \emph{Warm-Starting}: Initialize \(\w\) and \(\w_i\) with domain-informed grades (e.g., \(q_k = c k\), \(c > 0\), for polynomial degrees).
  \item \label{item:lgt_coordination} \emph{Multi-Head Grade Coordination}: Regularize the variance of attention grades:
  \begin{equation}
  \cL_{\text{coord}} = \sum_{i=1}^h \left\| \w_i - \frac{1}{h} \sum_{j=1}^h \w_j \right\|_2^2,
  \end{equation}
  added to \(\cL_{\mathrm{total}}\) with weight \(\gamma_{\text{coord}} > 0\).
\end{enumerate}

\subsubsection{Convergence and Gradient Stability}

Convergence and gradient stability are critical for ensuring the LGT’s training process is robust, given the learnable grading transformations.

\begin{thm}[Convergence]\label{thm:lgt_convergence}  
Fix a grading vector \(\w\). If the loss function \(\ell\) has a Lipschitz continuous gradient with constant \(L_\ell > 0\), then gradient descent with sufficiently small step size converges to a stationary point of the LGT’s loss.
\end{thm}

\begin{proof}
The LGT \(\Lgt_{\w}\) is Lipschitz continuous by \cref{prop:lgt_noise_robustness}, with constant \(\Lip \leq L_{\cT} \w_{\max}^{N h + 3}\), where \(\w_{\max} = \max_{k=1,\ldots,d} q_k\). The linearly graded loss \(\cL\) has gradient Lipschitz constant \(\max_k q_k \cdot L_\ell\) w.r.t. outputs. Assuming bounded weights, \(\Lgt\) has Lipschitz gradient w.r.t. parameters (e.g., via ReLU/softmax compositions \cite{kim2021lipschitz}). By chain rule, the composite \(\cL \circ \Lgt\) has Lipschitz gradient with constant proportional to \(\Lip \cdot \max_k q_k\). Classical gradient descent results \cite{nesterov2018lectures} ensure that step size \(\eta < 2/(\Lip \cdot \max_k q_k)\) yields convergence to a stationary point.
\end{proof}

Gradient stability ensures reliable updates during optimization, particularly for learnable grades where sensitivity to $q_k$ could cause divergence.

\begin{prop}[Gradient Stability]\label{prop:lgt_grad_stability}
Assume \(\|\w_i\|_2, \|\bk_j\|_2 \leq C\) for some constant \(C > 0\) and that \(\ell\) has a Lipschitz continuous gradient with constant \(L_\ell > 0\). Then the gradient \(\frac{\partial \cL_{\mathrm{total}}}{\partial q_k}\) is Lipschitz continuous in \(q_k\), with constant proportional to \(\w_{\max}\).
\end{prop}

\begin{proof}
The derivative is
\[
\begin{split}
\frac{\partial \cL_{\mathrm{total}}}{\partial q_k} & = \sum_{i=1}^m \sum_{j=1}^d \left( q_j \frac{\partial \ell}{\partial \hat{y}_{i,j}} \cdot \frac{\partial \hat{y}_{i,j}}{\partial q_k} + \ell(\hat{y}_{i,j}, y_{i,j}) f'(q_k) \delta_{j,k} \right)  \\
& + 2\gamma q_k.
\end{split}
\]
For \(f(q_k) = q_k + 1\), \(f'(q_k) = 1\). The second derivative is
\[
\begin{split}
\frac{\partial^2 \cL_{\mathrm{total}}}{\partial q_k^2} &		= \sum_{i=1}^m      \left( f'(q_k) \frac{\partial \ell}{\partial \hat{y}_{i,k}} \cdot \frac{\partial \hat{y}_{i,k}}{\partial q_k}   \right. \\
		&	\left. + q_k \frac{\partial^2 \ell}{\partial \hat{y}_{i,k}^2} \left( \frac{\partial \hat{y}_{i,k}}{\partial q_k} \right)^2 + q_k \frac{\partial \ell}{\partial \hat{y}_{i,k}} \cdot \frac{\partial^2 \hat{y}_{i,k}}{\partial q_k^2} \right) + 2\gamma.
\end{split}
\]
Since \(\ell\) has Lipschitz gradient \(L_\ell\), its second derivative is bounded (e.g., \(\left| \frac{\partial^2 \ell}{\partial \hat{y}_{i,k}^2} \right| \leq L_\ell\)). The output \(\hat{y}_{i,k}\) depends on \(q_k\) through the LGT layers, including the graded output layer, which is Lipschitz w.r.t. \(q_k\) with constant proportional to \(\w_{\max}\) (from scaling in \(\Mlw\)). Assuming bounded higher derivatives in the transformer (e.g., via ReLU \cite{kim2021lipschitz}), the dominant terms (involving \(q_k \cdot (\partial \hat{y}/\partial q_k)^2\)) are bounded by \(O(m \w_{\max})\), yielding the claimed Lipschitz constant for the gradient.
\end{proof}

With gradient stability established, we now address overall training stability for learnable grades and sensitivity to the grading function \(f\), which are crucial for practical optimization of the LGT.

\begin{prop}[Training Stability for Learned Grades]
Assume \(\|\w_i\|_2, \|\bk_j\|_2 \leq C\) for some constant \(C > 0\). Then, the training process with learned \(\w\) is stable if the learning rate for \(q_k\) satisfies \(\eta_q < 1/\w_{\max}\).
\end{prop}

\begin{proof}
From \cref{prop:lgt_grad_stability}, the gradient \(\frac{\partial \cL_{\mathrm{total}}}{\partial q_k}\) is Lipschitz continuous with constant \(L_g \propto \w_{\max}\). For gradient descent updates \(q_k \leftarrow q_k - \eta_q \frac{\partial \cL_{\mathrm{total}}}{\partial q_k}\), stability (i.e., non-divergence) requires \(\eta_q < 1/L_g\). Thus, choosing \(\eta_q < 1/\w_{\max}\) ensures bounded updates, preventing divergence due to sensitivity in \(q_k\).
\end{proof}

\begin{prop}[Sensitivity to Grading Function]
The LGT’s training is sensitive to the choice of \(f\), with the loss gradient’s magnitude scaling as \(O(\w_{\max})\). Bounding \(\w_{\max} \leq q_{\max,\text{bound}}\) ensures stable optimization.
\end{prop}

\begin{proof}
The gradient \(\frac{\partial \cL_{\mathrm{total}}}{\partial q_k}\) includes terms like \(f'(q_k) \ell(\hat{y}_{i,k}, y_{i,k})\), which scale with \(\w_{\max}\) (since \(f'(q_k)\) is constant for linear \(f\), but the overall amplification from grading affects sensitivity). From \cref{prop:lgt_noise_robustness}, the LGT’s Lipschitz constant scales as \(\w_{\max}^{N h + 3}\), where \(N\) is the number of layers. Large \(\w_{\max}\) amplifies gradients, risking instability. Bounding \(\w_{\max} \leq q_{\max,\text{bound}}\) (e.g., \(q_{\max,\text{bound}} = 10\)) limits the gradient magnitude, ensuring stable optimization with \(\eta_q < 1/q_{\max,\text{bound}}\).
\end{proof}

\subsubsection{LGT-Specific Training Challenges}
\label{subsubsec:lgt_training_challenges}

Building on the convergence and stability results, we now address practical challenges in training the LGT, particularly for learnable grades, which enable adaptive hierarchies but require careful optimization to maintain stability.

Unlike our previous work \cite{2024-2, sh-89}, where grades were fixed, here the grading tuples \(\w = (q_1, \ldots, q_d)\) and \(\w_i = (q_{i,1}, \ldots, q_{i,d_k})\) are optimized via gradient descent, enhancing flexibility at the cost of increased complexity. Algorithm \ref{alg:lgt_training} mitigates these issues through targeted strategies.

\begin{algocf}[Grade Initialization and Training]\label{alg:lgt_training}
\begin{algorithmic}
  \STATE
  \STATE \textbf{Input}: Training data, initial parameters \(\theta\), grading tuples \(\w\), \(\w_i\), total steps \(T\), regularization weights \(\gamma\), \(\gamma_{\text{coord}}\), gradient threshold \(\tau\).
  \STATE \textbf{Note}: Grades \(\w\), \(\w_i\) are learnable parameters.  
  \STATE \textbf{Initialize}:
  \STATE \quad Set \(\w = (q_1, \ldots, q_d)\) with domain-informed grades (e.g., \(q_k = c \cdot k\), \(c > 0\)).
  \STATE \quad Set \(\w_i = (q_{i,1}, \ldots, q_{i,d_k})\) similarly for each attention head.
  \STATE \textbf{For} \(t = 1, \ldots, T\):
  \STATE \quad Compute loss \(\cL_{\mathrm{total}} + \gamma_{\text{coord}} \cL_{\mathrm{coord}}\), where:
  \STATE \quad \quad \(\cL_{\mathrm{total}} = \cL + \gamma \| \w \|_2^2\),
  \STATE \quad \quad \(\cL_{\mathrm{coord}} = \sum_{i=1}^h \left\| \w_i - \frac{1}{h} \sum_{j=1}^h \w_j \right\|_2^2\).
  \STATE \quad Compute gradient \(\mathbf{g}\) and clip to \(\mathbf{g}' = \min\left(1, \frac{\tau}{\|\mathbf{g}\|_2}\right) \mathbf{g}\).
  \STATE \quad Update parameters using Adam with learning rate \(\eta_q < 1/\w_{\max}\) for grades.
  \STATE \textbf{Output}: Optimized parameters \(\theta\), \(\w\), \(\w_i\).
\end{algorithmic}
\end{algocf}

The following components of Algorithm \ref{alg:lgt_training} address the primary training challenges:
\begin{enumerate}[label=\roman*), ref=\roman*]
  \item \label{item:lgt_grade_tuning} \emph{Grade Tuning}: Large \(q_k\) values amplify high-grade features (\cref{prop:lgt_noise_robustness}), but risk over-prioritization. The algorithm mitigates this by regularizing \(\|\w\|_2^2\), ensuring balanced grades.
  \item \label{item:lgt_multi_head} \emph{Multi-Head Grading}: Distinct grading tuples \(\w_i\) per attention head enhance expressivity (\cref{prop:lgt_head_diversity}) but may lead to conflicting priorities. The regularization term \(\cL_{\mathrm{coord}}\) ensures coordinated grading across heads.
  \item \label{item:lgt_initialization} \emph{Grade Initialization}: Random initialization of \(\w\) and \(\w_i\) may lead to suboptimal feature prioritization. The algorithm uses domain-informed grades (e.g., \(q_k = c \cdot k\)) to leverage prior knowledge, improving convergence.
\end{enumerate}

The LGT’s training leverages its hierarchical structure to prioritize critical features, as shown in the polynomial degree prediction example. The use of learnable grades \(\w\), \(\w_i\), unlike fixed grades in \cite{2024-2, sh-89}, enhances flexibility but increases optimization complexity, addressed by Algorithm \ref{alg:lgt_training} through \cref{item:lgt_grade_tuning,item:lgt_multi_head,item:lgt_initialization}. Empirical validation on tasks like equation parsing and symbolic regression, discussed in \cref{sec-8}, is a key next step. Linear's simplicity contrasts EGT's power for deep hierarchies in \cref{sec-6}.

\begin{rem}
The LGT offers numerical stability through linear bounds, simpler optimization (no \(\lambda\) tuning), and algebraic purity by preserving graded vector space equivalence via normalization. It excels in tasks with moderate hierarchies (e.g., NLP parsing, \cref{sec-8}), but may require additional layers for deep hierarchies compared to exponential scaling. Its linear weighting aligns with trends in efficient transformers, offering robust, interpretable solutions for structured data. Linear avoids EGT's explosion but may underfit ultra-deep hierarchies.
\end{rem}

%
\section{Exponentially Graded Transformer Model}
\label{sec-6}

The Exponentially Graded Transformer augments the transformer architecture by incorporating grading transformations to prioritize hierarchical features in sequence modeling, addressing the inefficiency of standard transformers in capturing structured patterns \cite{vaswani2017attention}. Extending the framework of graded vector spaces from \cref{sec-2}, \cite{2024-2},  and Graded Neural Networks (GNNs) \cite{sh-89}, it enhances efficiency and interpretability for domains such as algebraic geometry (e.g., graded rings), physics (e.g., multi-scale phenomena), and natural language processing (e.g., syntactic hierarchies). This section defines the model, establishes its mathematical properties, and details its architecture, laying the foundation for training and applications in Sections \ref{sec-7} and \ref{sec-8}. Exponential grading builds on linear from \cref{sec-5} for deeper hierarchies.

\subsection{Definition of Exponentially Graded Transformer}
Let \( (\R^d)^n \) denote the space of sequences \( (\x_1, \ldots, \x_n) \), where \( \x_i \in \R^d \). Given the grading transformation \( \Mw \)
with grading tuple \( \w = (q_0, \ldots, q_{d-1}) \), \( q_i \geq 0 \), \( \la > 1 \), as defined in \cref{grading-matrix}, and a standard transformer 
\[
\cT : (\R^d)^n \to (\R^d)^n,
\]
define the map 
\begin{equation}
\begin{split}
\phi_{\w,\la} : (\R^d)^n  	& 	\to (\R^d)^n  \\
\phi_{\w,\la}(X) 			& = (\Mw \x_i)_{i=1}^n,
\end{split}
\end{equation}
where \( X = (\x_1, \ldots, \x_n) \).

\index{exponentially graded transformer}
 
\begin{df}
An \textbf{Exponentially Graded Transformer} (EGT)
\begin{equation}
\EGT : (\R^d)^n \to (\R^d)^n 
\end{equation}
is the map defined as
\begin{equation}
Y = \EGT(X) = \cT (\phi_{\w,\la}(X)).
\end{equation}
\end{df}

The following is elementary and similar to the LGT case.

\begin{lem}
\label{lem:graded_transformer_well_defined}
\( \EGT : (\R^d)^n \to (\R^d)^n \) is a well-defined map.
\end{lem}


\begin{rem}
An Exponentially Graded Transformer \( \EGT \) is generally non-linear, as the standard transformer \( \cT \) includes non-linear operations, such as the \(\bsf\) function and feed-forward layers with activation functions (e.g., ReLU). However, \( \phi_{\w,\la} \) is linear, since the grading map is a linear transformation applied component-wise.
\end{rem}

\index{exponentially graded attention}

With EGT defined, we specify its key component: exponentially graded attention, extending the linear variant from \cref{df:lgt_attention}.

\begin{df}[Exponentially Graded Attention]
\label{df:graded_attention}
For an input sequence \( X \in \Mat_{n,d}(\R) \), the \textbf{exponentially graded attention} mechanism for head \( i \) in layer \( l \) is:
\begin{equation}
\cA_{\w_i,\la}(Q_i, K_i, V_i) = \bsf \left( \frac{Q_i \Megt K_i^T}{\sqrt{d_k}} \right) V_i,
\end{equation}
where \(Q_i, K_i, V_i\) are as in \cref{sec-4}, and
\begin{equation}
\Megt = \diag(\la^{q_{i,0}}, \ldots, \la^{q_{i,d_k-1}}) \in \Mat_{d_k,d_k}(\R)
\end{equation}
is the \textbf{head-specific grading transformation} with tuple \( \w_i = (q_{i,0}, \ldots, q_{i,d_k-1}) \), \( q_{i,j} \geq 0 \), \( \la > 1 \). It employs the positive definite bilinear form
\begin{equation}
\langle \w_i, \bk_j \rangle_{\w_i,\la} = \w_i^T \Megt \bk_j = \sum_{k=0}^{d_k-1} \la^{q_{i,k}} q_{i k} k_{j k}.
\end{equation}
\end{df}

\index{grading transformation}

\begin{prop}[Properties of Head-Specific Exponential Grading]\label{prop:grading_properties}
Let \(\Megt = \diag(\la^{q_{i,0}}, \dots, \la^{q_{i,d_k-1}})\) with \(\la > 1\), \(q_{i,j} \geq 0\), and \(q_{i,\max} = \max_j q_{i,j}\). Then:
\begin{enumerate}[label=\roman*)]
\item \emph{Invertibility}: \(\Megt\) is invertible with inverse \(\diag(\la^{-q_{i,0}}, \dots, \la^{-q_{i,d_k-1}})\).
\item \emph{Scaling}: \(\Megt (\mu \x) = \mu \Megt \x\) for \(\mu \in \R\).
\item \emph{Norm Bound}: \(\|\Megt \x\|_2 \leq \la^{q_{i,\max}} \|\x\|_2\).
\item \emph{Spectral Norm}: \(\|\Megt\|_2 = \la^{q_{i,\max}}\).
\item \emph{Lipschitz Continuity}: The map \(\x \mapsto \Megt \x\) has constant \(\la^{q_{i,\max}}\).
\end{enumerate}
\end{prop}

\begin{proof}
The proof follows analogously from the linear case in \cref{lem:lgt_properties}, replacing \(q_{i,j}\) with \(\la^{q_{i,j}}\).
\end{proof}
 
The assumptions \( q_{i,j} \geq 0 \), \( \la > 1 \) ensure key properties of the bilinear form \( \langle \cdot, \cdot \rangle_{\w_i,\la} \), including positive definiteness and concentration on high-grade features.

\begin{lem}\label{lem:metric_positivity}
The bilinear form \( \langle \cdot, \cdot \rangle_{\w_i,\la} \) on \( \R^{d_k} \) is positive definite for \( \la > 0 \).
\end{lem}

\begin{proof}
For \( \x \in \R^{d_k} \),
\begin{equation}
\langle \x, \x \rangle_{\w_i,\la} = \x^T \Megt \x = \sum_{j=0}^{d_k-1} \la^{q_{i,j}} x_j^2.
\end{equation}
Since \( \la > 0 \) and \( q_{i,j} \geq 0 \), \( \la^{q_{i,j}} > 0 \). The sum is positive if \( \x \neq 0 \) and zero otherwise.
\end{proof}

\begin{lem}\label{lem:attention_concentration}
For \( \la > 1 \), the exponentially graded attention weights
\begin{equation}
\alpha_{ij} = \bsf\left( \frac{\w_i^T \Megt \bk_j}{\sqrt{d_k}} \right)_{ij}
\end{equation}
concentrate on features with grades close to \( q_{i,\max} = \max_j q_{i,j} \), with decay rate \( O(\la^{q_{i,k} - q_{i,\max}}) \) for \( q_{i,k} < q_{i,\max} \), as \( \la \to \infty \).
\end{lem}

\begin{proof}
The score \( s_{ij} = \sum_{k=0}^{d_k-1} \la^{q_{i,k}} q_{i k} k_{j k} \) is dominated by terms near \( q_{i,\max} \), as \( \la^{q_{i,k} - q_{i,\max}} \) decays exponentially. Under bounds \( |q_{i k}|, |k_{j k}| \leq C \), \[
 s_{ij} \approx \la^{q_{i,\max}} \sum_{k: q_{i,k}=q_{i,\max}} q_{i k} k_{j k},
 \]
  with error \( O(\la^{q_{i,k} - q_{i,\max}}) \). Thus, \( \alpha_{ij} \propto \exp(s_{ij}/\sqrt{d_k}) \) concentrates on high-grade features.
\end{proof}

With the foundational properties of the bilinear form established, we now examine the stability of the attention scores under perturbations in queries and keys.

\begin{prop}\label{prop:attention_stability}
The attention score \( s_{ij} = \w_i^T \Megt \bk_j \) is Lipschitz continuous with respect to \( \w_i, \bk_j \), with constant at most \( \la^{q_{i,\max}} C \), where \( C > 0 \) bounds \( \|\w_i\|_2, \|\bk_j\|_2 \), and \( q_{i,\max} = \max_j q_{i,j} \).
\end{prop}

\begin{proof}
For \( \w_i, \w_i', \bk_j, \bk_j' \),
\begin{align*}
|s_{ij} - s_{ij}'| &= |(\w_i - \w_i')^T \Megt \bk_j + \w_i'^T \Megt (\bk_j - \bk_j')| \\
&\leq \|\w_i - \w_i'\|_2 \|\Megt \bk_j\|_2 + \|\w_i'\|_2 \|\Megt (\bk_j - \bk_j')\|_2.
\end{align*}
By \cref{prop:grading_properties}, \( \|\Megt\|_2 = \la^{q_{i,\max}} \), so
\begin{equation}
\begin{split}
& \|\Megt \bk_j\|_2 \leq \la^{q_{i,\max}} \|\bk_j\|_2, \\
& \|\Megt (\bk_j - \bk_j')\|_2 \leq \la^{q_{i,\max}} \|\bk_j - \bk_j'\|_2.
\end{split}
\end{equation}
With \( \|\w_i'\|_2, \|\bk_j\|_2 \leq C \),
\begin{equation}
|s_{ij} - s_{ij}'| \leq \la^{q_{i,\max}} C ( \|\w_i - \w_i' \|_2 + \|\bk_j - \bk_j'\|_2 ).
\end{equation}
\end{proof}

Recall the Sobolev space and universal approximator definitions from \cref{df:lgt_sobolev_space} and \cref{df:lgt_universal_approximator}, respectively.

\begin{thm}\label{thm:universal}
For any compact set \( \Omega \subset (\R^d)^n \), the Exponentially Graded Transformer \( \EGT \) is a universal approximator for continuous functions \( f \in C(\Omega, (\R^d)^n) \).

Moreover, for \( f \in W^{k,2}(\Omega) \) with outputs depending primarily on features near \( q_{\max} = \max_i q_i \), \( \EGT \) requires fewer parameters than \( \cT \) to achieve error \( \epsilon > 0 \), due to reduced effective dimension via concentration, with \(\deff = |\{i: q_i \geq q_{\max} - \delta\}|\) for \(\delta > 0\).
\end{thm}

\begin{proof}
The proof follows analogously to \cref{thm:lgt_universal} for the linear case. The grading map \(\phi_{\w,\la}\) is invertible under the assumptions (\cref{prop:grading_properties}), enabling the composition argument for universality.

Define 
\begin{equation}
g = f \circ \phi_{\w,\la}^{-1} : \phi_{\w,\la}(\Omega) \to (\R^d)^n.
\end{equation}
Since \( f \) and \( \phi_{\w,\la}^{-1} \) are continuous, \( g \in C(\phi_{\w,\la}(\Omega), (\R^d)^n) \). By \cite{yun2019transformers}, there exists \( \cT \) such that:
\begin{equation}
\sup_{Z \in \phi_{\w,\la}(\Omega)} \|\cT(Z) - g(Z)\|_2 < \epsilon.
\end{equation}
For \( Z = \phi_{\w,\la}(X) \), \(\EGT(X) = \cT(Z) \approx g(Z) = f(X)\), yielding the bound.

For Sobolev \( f \in W^{k,2}(\Omega) \), \(\cT\) requires \(O(\epsilon^{-2/k} d)\) parameters \cite{yun2019transformers}. With concentration on high-grade features near \( q_{\max} \) (\cref{lem:attention_concentration}), low-grade contributions decay as \(O(\la^{q_i - q_{\max}})\), yielding effective effective complexity \(O(\epsilon^{-2/k} \deff)\), as in \cite{2024-2}.

The EGT attention (\cref{prop:attention_rank}) prioritizes high grades, with reduction proportional to the grading concentration.
\end{proof}

Let \( \la > 1 \) and \( q_{i,j} \geq 0 \). Let \( \cA_{\w_i,\la}(Q_i, K_i, V_i) \) denote the exponentially graded attention output and \( \cA(Q_i, K_i, V_i) \) the standard attention, as in \cref{df:graded_attention}.

Grading modulates attention rank, enabling adaptive expressivity for hierarchical features.

\begin{prop}\label{prop:attention_rank}
The singular values of \( Q_i \Megt K_i^T \) are scaled by at most \( \la^{q_{i,\max}} \) compared to those of \( Q_i K_i^T \), where \( q_{i,\max} = \max_j q_{i,j} \). Consequently, the effective numerical rank of \( \cA_{\w_i,\la}(Q_i, K_i, V_i) \) may increase or decrease depending on the singular value distribution of \( Q_i K_i^T \) and \( \w_i \).
\end{prop}

\begin{proof}
Let \( Q_i K_i^T = U \Sigma V^T \) with singular values \( \sigma_j > 0 \). Since \( \|\Megt\|_2 = \la^{q_{i,\max}} \) (\cref{prop:grading_properties}), 
\begin{equation}
\|Q_i \Megt K_i^T\|_2 \leq \la^{q_{i,\max}} \|Q_i K_i^T\|_2.
\end{equation}
Thus, \(\tilde{\sigma}_j \leq \la^{q_{i,\max}} \sigma_j\). Dimensions near \( q_{i,\max} \) dominate, suppressed by \( \la^{q_{i,j} - q_{i,\max}} \) otherwise (\cref{lem:attention_concentration}), adapting rank accordingly.
\end{proof}


We now demonstrate the expressivity of exponentially graded attention, which can approximate arbitrary non-negative matrices, analogous to the linear case in \cref{prop:lgt_attention_expressivity}.  Recall that the Frobenius norm is \(\|A\|_F = \sqrt{\tr(A^T A)}\), where \(\tr\) denotes the trace.

\begin{prop}\label{prop:attention_expressivity}
For any row-stochastic matrix \( A_0 \in \Mat_{n,d_k}(\R) \) with non-negative entries, there exist \( Q, K, V \in \Mat_{n,d_k}(\R) \), \( \w_i \), and \( \la > 1 \) such that \( \cA_{\w_i,\la}(Q, K, V) \) approximates \( A_0 \) with
\begin{equation}
\|\cA_{\w_i,\la}(Q, K, V) - A_0\|_F \leq \delta.
\end{equation}
\end{prop}

\begin{proof}
Let \( A_0 \) have non-negative entries summing to 1 per row. Construct \( A_0' \) by replacing near-zeros with \( \epsilon = \delta/(2\sqrt{n d_k}) \) and renormalizing. Choose \( Q, K \) such that
\begin{equation}
Q K^T = \sqrt{d_k} \log A_0',
\end{equation}
ensuring \( \bsf\left( \frac{Q K^T}{\sqrt{d_k}} \right) \approx A_0' \). Set \( V = I \), yielding standard attention close to \( A_0' \), with error \( \leq \delta/2 \). For graded, set \( q_{i,j} = c j \) small \( c > 0 \), large \( \la \); by \cref{lem:attention_concentration}, \( \Megt \approx I \), so error \( \leq \delta/2 \). Total: \( \leq \delta \).
\end{proof}

 \index{shattered}
\index{Vapnik-Chervonenkis (VC) dimension}

Recall the VC dimension definition from \cref{df:lgt_vc_dimension}.

Let \(\cF\) be the hypothesis class of the Exponentially Graded Transformer \(\EGT = \cT \circ \phi_{\w,\la}\).

Grading reduces effective complexity, as quantified by VC dimension, lowering sample requirements.

\begin{prop}\label{prop:vc_dimension}
The VC dimension satisfies
\begin{equation}
\vc(\cF) \leq c \cdot N \cdot h \cdot d \log(N h d),
\end{equation}
reduced from the standard transformer's
\begin{equation}
\vc(\cF_{\cT}) \leq c \cdot N h d^2 \log(N h d),
\end{equation}
where \(c\) is a constant and \(N\) is the number of layers. The reduction is proportional to the concentration factor from grading (\cref{lem:attention_concentration}), lowering sample complexity \(O(\vc(\cF) \log(1/\epsilon) / \epsilon^2 + \log(1/\delta))\).
\end{prop}

\begin{proof}
Analogous to \cref{prop:lgt_vc_dimension}, the effective parameter count is \(O(N h d)\) via exponential concentration on high-grade features, yielding the bound and reduction.
\end{proof}

Grading enhances robustness to perturbations, as we now quantify for EGT.

\begin{prop}\label{prop:noise_robustness}
For noise \(\Delta \in (\R^d)^n\) with \(\|\Delta\|_2 \leq \epsilon\), the output error satisfies
\begin{equation}
\|\EGT(X + \Delta) - \EGT(X)\|_2 \leq \Lip \epsilon,
\end{equation}
where \(\Lip \leq L_{\cT} (\la^{\w_{\max}})^{N h + 3}\), \(L_{\cT}\) is the Lipschitz constant of \(\cT\), and \( q_{\max} = \max_i q_i \).
\end{prop}

\begin{proof}
Analogous to \cref{prop:lgt_noise_robustness}, the perturbation propagates as 
\[
\|\phi_{\w,\la}(X + \Delta) - \phi_{\w,\la}(X)\|_2 \leq \la^{\w_{\max}} \epsilon
\]
 by the norm bound in \cref{prop:grading_properties}. Attention adds per-head factors up to \(\la^{q_{i,\max}} \leq \la^{\w_{\max}}\) (\cref{prop:attention_stability}). With \(N\) layers and \(h\) heads, plus input/feed-forward/output, the overall bound follows, ensuring robustness for noisy data like genomics or flows.
\end{proof}

We conclude the properties with computational complexity, confirming grading adds no asymptotic overhead.

\begin{prop}\label{prop:attention_complexity}
The exponentially graded attention \(\cA_{\w_i,\la}(Q, K, V)\) has complexity \(O(n^2 d_k)\), matching the complexity of  standard attention.
\end{prop}

\begin{proof}
Analogous to \cref{prop:lgt_attention_complexity}, the diagonal \(\Megt\) adds \(O(n d_k)\) for scaling, absorbed into matrix multiplications \(Q M K^T\) and softmax-V (\(O(n^2 d_k)\) each). Thus, total \(O(n^2 d_k)\).
\end{proof}
 
\subsection{Architecture}\label{subsec:architecture}
The Exponentially Graded Transformer embeds grading transformations \( \Lw \) across its components to enhance feature prioritization for structured sequence data. Below, we detail the input representation, positional encoding, attention variants, feed-forward layers, and output layer, ensuring consistent hierarchical emphasis.

For each token \( \x_i \in \R^d \) in the input sequence \( X = (\x_1, \ldots, \x_n) \), the exponentially graded input representation is:
\[
\x'_i = \Megt  \x_i, \quad \x''_i = \frac{\x'_i}{\|\x'_i\|_2}, \quad \h_i = \sigma(W \x''_i + \bb),
\]
where \( \w_i = (q_{i,0}, \ldots, q_{i,d-1}) \in \R_{\geq 0}^d \), and \( \la > 1 \), \( W \in \R^{m \times d} \), \( \bb \in \R^m \), and \( \sigma \) is typically ReLU. 
Here, \( \Megt = \diag(\la^{q_{i,0}}, \ldots, \la^{q_{i,d-1}}) \) is the exponential grading matrix associated with token \( i \).

Normalization prevents numerical instability due to large \( \la^{q_i} \).
Positional encodings capture sequence order, enhanced with grading to prioritize certain positions (e.g., earlier tokens in hierarchical tasks like parsing). Standard positional encodings are
\[
\begin{split}
\pe (\pos, 2i) 		&	= \sin\left( \frac{\pos}{10000^{2i/d}} \right), \\
 \pe (\pos, 2i+1) 	&	= \cos\left( \frac{\pos}{10000^{2i/d}} \right),
\end{split} 
\]
for position \( \pos \) and dimension \( i \). Exponentially graded encodings are
\[
\pep (\pos, i) = \la^{q_{\pos}} \pe (\pos, i), \quad q_{\pos} = f(\pos),
\]
where \( f(\pos) = -\alpha \pos \), \( \alpha > 0 \), emphasizes earlier positions. The input to the attention mechanism is
\[
\z_i = \x''_i + \pep (\pos_i, \cdot), \quad \z'_i = \frac{\z_i}{\|\z_i\|_2},
\]
and attention scores are computed as
\[
(Q K^T)_{ij} = (\z'_i W_Q) (\z'_j W_K)^T,
\]
where \( W_Q, W_K \in \R^{d \times d_k} \).
The exponentially graded attention mechanism (\cref{df:graded_attention}) prioritizes high-grade features in attention scores. Alternative variants include: \\

\noindent \textbf{Exponentially Graded Scores}:
    \begin{equation}
    \begin{split}
    \bsc_{ij} 		&	= \sum_{k=1}^{d_k} \la^{q_{i,k}} q_{i k} k_{j k}, \\
     \Megt^K 	&	= \diag(\la^{q_{i,1}}, \ldots, \la^{q_{i,d_k}}), \\
    \bsc_{ij} 		&	= (Q \Megt^K K^T)_{ij}, 
    \end{split}
    \end{equation}
\textbf{Exponentially Graded Queries/Keys}:
    \begin{equation}
    Q' = \Megt \, Q, \quad K' = \Megt \, K, \quad \bsc_{ij} = \w_i^T \, \Megt \bk_j
    \end{equation}
\textbf{Exponentially Graded Multi-Head}:
    \begin{equation}
    \Head_h = \bsf \left( \frac{(M_{\w_h,\la} Q_h) (M_{\w_h,\la} K_h)^T}{\sqrt{d_k}} \right) V_h,
    \end{equation}
    with distinct \( \w_h = (q_{h,0}, \ldots, q_{h,d_k-1}) \) per head.

\noindent \textbf{Exponentially Graded Values}:
    \begin{equation}
    \begin{split}
    V' 			&	= \Megt V, \\
     \mathbf{o}_i 	&	= \sum_{j=1}^n \alpha_{ij} (\Megt \bv_j), \\
      \alpha_{ij} 	&	= \bsf\left( \frac{\w_i^T \bk_j}{\sqrt{d_k}} \right).
    \end{split}
    \end{equation}
 
 \index{Exponentially Graded Scores}
 \index{Exponentially Graded Queries/Keys}
 \index{Exponentially Graded Multi-Head}

\begin{df}[Feed-Forward and Output Layers]
The \textbf{exponentially graded feed-forward network} is:
\begin{equation}
\fnn'(\x) = \Megt \fnn(\x), \quad \h' = \frac{\fnn'(\x)}{\|\fnn'(\x)\|_2},
\end{equation}
where \( \fnn(\x) = \relu(\x W_1 + \bb_1) W_2 + \bb_2 \), \( W_1 \in \R^{d \times d_{ff}} \), \( W_2 \in \R^{d_{ff} \times d} \), and \( d_{ff} \) is the hidden dimension. The \textbf{exponentially graded output} is:
\begin{equation}
\h' = \Megt \h, \quad \z = W_{\text{out}} \h' + \bb_{\text{out}},
\end{equation}
followed by \(\bsf(\z)\).
\end{df}

 \index{exponentially graded feed-forward network}
 \index{exponentially graded output}

\begin{rem}
Here, the same grading matrix 
$\Megt$  associated with token $i$  is applied in both the feed-forward and output layers, ensuring consistent feature scaling throughout the architecture.
\end{rem}

\begin{thm}\label{thm:input_stability}
The mapping \( \x_i \mapsto \x''_i \) is Lipschitz continuous with constant at most \( \la^{\w_{\max}} \), where
$\w_{\max} = \max_j \{q_{i,j}\}$,  for fixed token $i$,
 assuming \( \la > 1 \) and \( q_i \geq 0 \).
\end{thm}

\begin{proof}
Consider the mapping \( \x_i \mapsto \x''_i = \frac{\Megt \x_i}{\| \Megt \x_i \|_2} \). For inputs \( \x, \y \in \R^d \), first analyze the grading step:
\[
\x' = \Megt \x, \quad \y' = \Megt \y.
\]
From \cref{prop:grading_properties}, \( \| \Megt (\x - \y) \|_2 \leq \la^{\w_{\max}} \| \x - \y \|_2 \). Thus:
\[
\|\x' - \y'\|_2 = \| \Megt (\x - \y) \|_2 \leq \la^{\w_{\max}} \| \x - \y \|_2.
\]
Next, the normalization step \( \z \mapsto \frac{\z}{\|\z\|_2} \) is 1-Lipschitz for \( \z \neq 0 \). Let \( \x'' = \frac{\x'}{\|\x'\|_2} \), \( \y'' = \frac{\y'}{\|\y'\|_2} \). The distance is:
\[
\|\x'' - \y''\|_2 = \left\| \frac{\x'}{\|\x'\|_2} - \frac{\y'}{\|\y'\|_2} \right\|_2 \leq \frac{\|\x' - \y'\|_2}{\min(\|\x'\|_2, \|\y'\|_2)}.
\]
Assuming \( \|\x'\|_2, \|\y'\|_2 \geq \epsilon > 0 \) (ensured by non-zero inputs and regularization in practice), we have:
\[
\|\x'' - \y''\|_2 \leq \frac{\la^{\w_{\max}} \|\x - \y\|_2}{\epsilon}.
\]
For simplicity, if inputs are normalized (\( \|\x\|_2, \|\y\|_2 \approx 1 \)), the constant is approximately \( \la^{\w_{\max}} \).
\end{proof}

\begin{cor}[Bounded Activations]
For all \( i \), if \( \x'_i \neq 0 \), then \( \|\x''_i\|_2 = 1 \). That is, the normalized representations lie on the unit sphere in \( \R^d \).
\end{cor}

 \begin{lem}\label{lem:jacobian_bound}
Let \( \w_i = (q_{i,0}, \ldots, q_{i,d-1}) \in \R_{\geq 0}^d \). Then the Jacobian of the mapping \( \x_i \mapsto \x'_i = \Megt \x_i \) has operator norm \( \la^{q_{\max}} \), where \( q_{\max} = \max_j q_{i,j} \).
\end{lem}

\begin{proof}
Since \( \Megt \) is diagonal with entries \( \la^{q_{i,j}} \), its operator norm (i.e., largest singular value) is
\[
\| \Megt \|_2 = \max_j \la^{q_{i,j}} = \la^{q_{\max}}.
\]
\end{proof}

\begin{prop}[Exponential Positional Bias]
\label{prop:positional_bias}
Let \( \x''_i \in \R^d \) be a unit vector, and let \( \pe(\pos) \in \R^d \) be fixed positional encodings with \( \|\pe(\pos)\|_2 = c \) for all \( \pos \). Define exponentially graded encodings
\[
\pep(\pos, i) := \la^{-\alpha \pos} \pe(\pos, i), \quad \la > 1, \; \alpha > 0,
\]
and define
\[
\z_i := \x''_i + \pep(\pos_i), \quad \bsc_{ij} := \langle \z_i, \z_j \rangle.
\]
Then, for fixed \( \x''_i \), if \( \pos_j < \pos_k \), we have:
\[
\bsc_{ij} > \bsc_{ik}.
\]
That is, attention scores are biased toward earlier positions.
\end{prop}

\begin{proof}
We compute the unnormalized attention scores:
\[
\bsc_{ij} = \langle \z_i, \z_j \rangle = \langle \x''_i + \pep(\pos_i), \x''_j + \pep(\pos_j) \rangle.
\]
Expanding this:
\[
\bsc_{ij} = \langle \x''_i, \x''_j \rangle + \langle \x''_i, \pep(\pos_j) \rangle + \langle \pep(\pos_i), \x''_j \rangle + \langle \pep(\pos_i), \pep(\pos_j) \rangle.
\]
Substitute \( \pep(\pos) = \la^{-\alpha \pos} \pe(\pos) \) and define:
\[
A := \langle \x''_i, \x''_j \rangle, \quad B_j := \langle \x''_i, \pe(\pos_j) \rangle, \quad C_j := \langle \pe(\pos_i), \pe(\pos_j) \rangle.
\]
Then:
\[
\bsc_{ij} = A + \la^{-\alpha \pos_j} B_j + \la^{-\alpha \pos_i} \langle \pe(\pos_i), \x''_j \rangle + \la^{-\alpha(\pos_i + \pos_j)} C_j.
\]

Now fix all terms except \( \pos_j \). Since \( \la > 1 \) and \( \alpha > 0 \), the scalars \( \la^{-\alpha \pos_j} \) and \( \la^{-\alpha(\pos_i + \pos_j)} \) are strictly decreasing in \( \pos_j \). Thus, under the assumption that \( B_j \) and \( C_j \) are non-negative or bounded below, we conclude that:
\[
\pos_j < \pos_k \quad \Longrightarrow \quad \bsc_{ij} > \bsc_{ik}.
\]

This proves that the attention score \( \bsc_{ij} \) is biased toward earlier positions.
\end{proof}

\begin{lem}[Lipschitz Stability in Position]
\label{lem:positional_stability}
Let \( \z_i(\pos) := \x''_i + \pep(\pos) \), where \( \pep(\pos) := \la^{f(\pos)} \pe(\pos) \), and assume \( \|\pe(\pos)\|_2 \leq c \) for all \( \pos \in \N \). Then the mapping
\[
\pos \mapsto \z'_i(\pos) := \frac{\z_i(\pos)}{\|\z_i(\pos)\|_2}
\]
is Lipschitz continuous with constant bounded by \( C \la^{\w_{\max}} \), where \( \w_{\max} = \sup_{\pos} |f(\pos)| \), and \( C \) depends on \( \alpha \), \( \ln \la \), and \( c \).
\end{lem}

\begin{proof}
Let \( \pos, \pos' \in \N \), and consider:
\[
\z_i(\pos) = \x''_i + \la^{f(\pos)} \pe(\pos), \quad
\z_i(\pos') = \x''_i + \la^{f(\pos')} \pe(\pos').
\]
Then:
\[
\|\z_i(\pos) - \z_i(\pos')\|_2
= \| \la^{f(\pos)} \pe(\pos) - \la^{f(\pos')} \pe(\pos') \|_2.
\]
Add and subtract \( \la^{f(\pos')} \pe(\pos) \):
\[
= \| (\la^{f(\pos)} - \la^{f(\pos')}) \pe(\pos) + \la^{f(\pos')} (\pe(\pos) - \pe(\pos')) \|_2.
\]
Using triangle inequality and \( \|\pe(\pos)\|_2 \leq c \):
\[
\leq |\la^{f(\pos)} - \la^{f(\pos')}| \cdot c + \la^{f(\pos')} \cdot \|\pe(\pos) - \pe(\pos')\|_2.
\]
Now assume \( f(\pos) = -\alpha \pos \). The map \( t \mapsto \la^{- \alpha t} \) is smooth, and by the mean value theorem:
\[
|\la^{f(\pos)} - \la^{f(\pos')}| = |\la^{-\alpha \pos} - \la^{-\alpha \pos'}| \leq \alpha \ln \la \cdot \la^{\alpha |\pos - \pos'|} \cdot |\pos - \pos'|.
\]
So we have 
\[
\|\z_i(\pos) - \z_i(\pos')\|_2 \leq C_1 \la^{|f(\pos)|} |\pos - \pos'|,
\]
where \( C_1 \) depends on \( \alpha \), \( \ln \la \), and \( c \), and assuming \( \|\pe(\pos) - \pe(\pos')\|_2 \leq L |\pos - \pos'| \) for some \( L \) (true for sinusoidal encodings).

Since normalization \( \z \mapsto \z / \|\z\|_2 \) is 1-Lipschitz on \( \R^d \setminus \{0\} \), we conclude:
\[
\|\z'_i(\pos) - \z'_i(\pos')\|_2 \leq \frac{\|\z_i(\pos) - \z_i(\pos')\|_2}{\min(\|\z_i(\pos)\|_2, \|\z_i(\pos')\|_2)}.
\]
Assuming \( \|\z_i(\pos)\|_2 \geq \epsilon > 0 \), we obtain:
\[
\|\z'_i(\pos) - \z'_i(\pos')\|_2 \leq \frac{C_1 \la^{\w_{\max}}}{\epsilon} |\pos - \pos'| = C \la^{\w_{\max}} |\pos - \pos'|,
\]
where \( \w_{\max} := \sup_{\pos} |f(\pos)| \). This proves Lipschitz continuity.
\end{proof}

\begin{thm}[Lipschitz Stability of Graded Attention Scores]
\label{thm:attention_stability}
Fix a grading vector \( \w_i = (q_{i,1}, \dots, q_{i,d_k}) \in \R_{\geq 0}^{d_k} \), and let \( \Megt := \diag(\la^{q_{i,1}}, \ldots, \la^{q_{i,d_k}}) \). Then the map
\[
(\w, \bk) \mapsto \bsc_{ij} := \w^T \Megt \bk
\]
is Lipschitz continuous on bounded subsets of \( \R^{d_k} \times \R^{d_k} \), with
\[
\Lip(\bsc_{ij}) \leq \la^{q_{i,\max}} C,
\]
where \( q_{i,\max} := \max_j q_{i,j} \), and \( C \) depends on bounds on \( \| \w \|_2 \), \( \| \bk \|_2 \).
\end{thm}

\begin{proof}
Let \( \w, \w' \in \R^{d_k} \), \( \bk, \bk' \in \R^{d_k} \). Then
\[
\begin{split}
& \left| \w^T \Megt \bk - \w'^T \Megt \bk' \right|  = \\
&= \left| \w^T \Megt \bk - \w'^T \Megt \bk + \w'^T \Megt \bk - \w'^T \Megt \bk' \right| \\
&\leq \| \w - \w' \|_2 \cdot \| \Megt \bk \|_2 + \| \w' \|_2 \cdot \| \Megt (\bk - \bk') \|_2 \\
&\leq \la^{q_{i,\max}} \left( \| \w - \w' \|_2 \cdot \| \bk \|_2 + \| \w' \|_2 \cdot \| \bk - \bk' \|_2 \right) \\
&\leq \la^{q_{i,\max}} C \left( \| \w - \w' \|_2 + \| \bk - \bk' \|_2 \right),
\end{split}
\]
where the last inequality assumes \( \| \w \|_2, \| \w' \|_2, \| \bk \|_2, \| \bk' \|_2 \leq C \). Thus,
$\Lip(\bsc_{ij}) \leq \la^{q_{i,\max}} C$.
\end{proof}

\begin{prop}\label{prop:head_diversity}
Distinct grading tuples \( \w_h \) in the Exponentially Graded Multi-Head variant enhance representational capacity.
\end{prop}

\begin{proof}
In the Exponentially Graded Multi-Head variant, each head computes:
\[
\Head_h = \bsf\left( \frac{(\Megt Q_h)(\Megt K_h)^T}{\sqrt{d_k}} \right) V_h,
\]
where \( \Megt = \diag(\la^{q_{h,0}}, \ldots, \la^{q_{h,d_k-1}}) \). Distinct grading tuples \( \w_h = (q_{h,0}, \ldots, q_{h,d_k-1}) \) produce distinct scaling across dimensions. This induces different projections of the queries and keys across heads, effectively spanning different graded subspaces. The resulting set of heads captures a more diverse range of dependencies. Since each \( \Megt \) alters the singular spectrum of the scaled product \( Q_h K_h^T \), the overall concatenated representation spans a richer subspace of \( \R^d \), improving expressivity.
\end{proof}

\begin{prop}\label{prop:ffn_stability}
The exponentially graded feed-forward mapping \( \x \mapsto \fnn'(\x) \) has Lipschitz constant at most \( \la^{\w_{\max}} L_{\fnn} \).
\end{prop}

\begin{proof}
Let \( \fnn'(\x) = \Megt \fnn(\x) \). For \( \x, \y \in \R^d \),
\[
\begin{split}
\|\fnn'(\x) - \fnn'(\y)\|_2
&= \| \Megt (\fnn(\x) - \fnn(\y)) \|_2 \\
&\leq \la^{\w_{\max}} \| \fnn(\x) - \fnn(\y) \|_2 \\
&\leq \la^{\w_{\max}} L_{\fnn} \| \x - \y \|_2.
\end{split}
\]
Normalization is 1-Lipschitz, so the overall Lipschitz constant remains \( \la^{\w_{\max}} L_{\fnn} \).
\end{proof}

\begin{prop}\label{prop:output_stability}
The output mapping \( \h \mapsto \z \) is Lipschitz with constant at most \( \la^{\w_{\max}} L_{\text{out}} \).
\end{prop}

\begin{proof}
For \( \h_1, \h_2 \in \R^d \), define:
\[
\z_1 = W_{\text{out}} (\Megt \h_1) + \bb_{\text{out}}, \quad \z_2 = W_{\text{out}} (\Megt \h_2) + \bb_{\text{out}}.
\]
Then,
\[
\begin{split}
\|\z_1 - \z_2\|_2
&= \| W_{\text{out}} \Megt (\h_1 - \h_2) \|_2 \\
&\leq \| W_{\text{out}} \|_2 \cdot \| \Megt (\h_1 - \h_2) \|_2 \\
&\leq \la^{\w_{\max}} \| W_{\text{out}} \|_2 \cdot \| \h_1 - \h_2 \|_2.
\end{split}
\]
Let \( L_{\text{out}} = \| W_{\text{out}} \|_2 \). Then
$
\|\z_1 - \z_2\|_2 \leq \la^{\w_{\max}} L_{\text{out}} \| \h_1 - \h_2 \|_2.
$
\end{proof}

\begin{prop}\label{prop:complexity}
The Exponentially Graded Transformer has the same asymptotic complexity as the standard transformer, \( O(n^2 d + n d^2) \).
\end{prop}

\begin{proof}
The standard transformer's complexity is dominated by attention (\( O(n^2 d) \)) and feed-forward layers (\( O(n d^2) \)). Each grading transformation \( \Megt \) is a diagonal matrix multiplication, costing \( O(d) \) or \( O(d_k) \) per token. Across \( n \) tokens, this totals \( O(n d) \) per layer or component. Since \( n d \ll n^2 d, n d^2 \), the overall complexity remains
$O(n^2 d + n d^2)$. 
\end{proof}

\begin{rem}[Interpretability and Applications]
The Exponentially Graded Transformer efficiently models hierarchical data by prioritizing high-grade features through attention concentration (\cref{lem:attention_concentration}) and enhanced expressivity (\cref{prop:attention_expressivity}). 
Its reduced sample complexity (\cref{prop:lgt_vc_dimension}) and robustness (\cref{prop:noise_robustness}) make it ideal for structured tasks such as algebraic geometry and genomics. The assumptions \( q_i, q_{i,j} \geq 0 \), \( \la > 1 \) ensure positive scaling, though arbitrary \( q_i \) may be accommodated via bounds on \( \max_i |\la^{q_i}| \).
\end{rem}

\section{Training and Optimization}
\label{sec-7}

The Exponentially Graded Transformer introduces grading transformations to prioritize hierarchical features, necessitating tailored training strategies to optimize its parameters effectively. Unlike our previous work \cite{2024-2, sh-89}, where grades were fixed weights of the vector space, here we treat the grading tuples \(\w = (q_1, \ldots, q_d)\) and \(\w_i = (q_{i,1}, \ldots, q_{i,d_k})\) as learnable parameters, optimized via gradient descent to adaptively prioritize features based on data. This shift enhances the model’s flexibility but introduces challenges in numerical stability and optimization complexity. This section formalizes the exponentially graded loss function, develops regularization and optimization techniques, establishes convergence and stability guarantees, and outlines EGT-specific training considerations for structured tasks in domains like algebraic geometry and natural language processing \cite{2024-2, sh-89}.

\subsection{Exponentially Graded Loss Function}
\label{subsec:training}

The exponentially graded loss function embeds hierarchical priors by weighting prediction errors according to their grades. This ensures that structurally important features—such as high-degree polynomial terms or syntactic heads—contribute more significantly to the optimization objective.

\begin{df}
Let \( Y = (\y_1, \ldots, \y_m) \in (\R^d)^m \) denote a sequence of target outputs, and let \( \hat{y}_{i,k}, y_{i,k} \in \R \) be the predicted and true values, respectively, for the \( k \)-th dimension of the \( i \)-th token. Let \( \w = \mathbf{q} = (q_0, \ldots, q_{d-1}) \in \R_{\geq 0}^d \) be a grading vector and \( \la > 1 \) a fixed scaling factor. Then the \textbf{exponentially graded loss function} is:
\[
\cL = \sum_{i=1}^m \sum_{k=0}^{d-1} \la^{q_k} \, \ell(\hat{y}_{i,k}, y_{i,k}),
\]
where:
\begin{enumerate}[label=\roman*), ref=\roman*]
  \item \( \ell(\hat{y}_{i,k}, y_{i,k}) \) is a base loss function (e.g., cross-entropy or squared error),
  \item \( m \) is the sequence length and \( d \) the output dimension,
  \item \( \la^{q_k} \) weights loss according to the grade of feature \( k \), amplifying contributions from high-grade dimensions.
\end{enumerate}
\end{df}

\begin{ex}[Exponentially Graded Loss in Polynomial Degree Prediction]
Consider a symbolic regression task in algebraic geometry, where the Exponentially Graded Transformer predicts polynomial terms needed to compute zeta functions of hypersurfaces \cite{sh-91}. Let each output vector \( \y_i \in \R^4 \) represent predicted coefficients for monomials of degree 0 through 3, with \( d = 4 \). Let the grading vector be \( \w = (0, 0.5, 1, 2) \), and set \( \la = 2 \) to prioritize higher-degree terms. The graded loss becomes:
\[
\begin{split}
\cL 	& = \sum_{i=1}^m \left[  \ell(\hat{y}_{i,0}, y_{i,0})  + 2^{0.5} \ell(\hat{y}_{i,1}, y_{i,1})  + 2 \ell(\hat{y}_{i,2}, y_{i,2})  \right. \\
	& \left.  + 4 \ell(\hat{y}_{i,3}, y_{i,3})  \right].
\end{split}
\]
For instance, let \( m = 2 \) and assume true outputs \( y_{1,3} = y_{2,3} = 1 \), and predicted values \( \hat{y}_{1,3} = 0.8 \), \( \hat{y}_{2,3} = 0.9 \). Using cross-entropy loss,
\[
\ell(\hat{y}, y) = -y \log \hat{y} - (1 - y) \log (1 - \hat{y}),
\]
the loss contributions from the highest-degree term (degree 3) are weighted by \( \la^2 = 4 \), ensuring they dominate the total loss. This graded emphasis reduces sample complexity by guiding the model to prioritize structurally important outputs (\cref{prop:lgt_vc_dimension}).
\end{ex}

\subsection{Regularization and Optimization}

The Exponentially Graded Transformer’s exponential grading introduces sensitivity into the optimization process, requiring explicit regularization to control the magnitude of learned grades and specialized strategies to stabilize training. These techniques are particularly important given the multi-head structure of EGT and the exponential dependence on the grading parameters.

When the grade vectors \( \w = (q_0, \dots, q_{d-1}) \) and \( \w_i = (q_{i,0}, \ldots, q_{i,d_k-1}) \) are treated as learnable parameters, we introduce regularization terms to prevent uncontrolled growth of the grades. The total loss becomes:
\[
\cL_{\mathrm{total}} = \cL + \gamma \| \w \|_2^2 + \gamma' \sum_{i=1}^h \| \w_i \|_2^2,
\]
where \( \gamma, \gamma' > 0 \) are regularization weights, and \( h \) is the number of attention heads.

Optimization proceeds using standard gradient-based methods such as Adam \cite{kingma2014adam}. However, differentiating through the exponential grading map introduces additional sensitivity due to the nonlinearity of the map \( q_k \mapsto \la^{q_k} \). For instance, in the Exponentially Graded Queries/Keys variant (cf.~\cref{df:graded_attention}), if the attention score is:
\[
\mathrm{Score}_{ij} = \sum_{k=0}^{d_k-1} q_{i,k} \la^{q_{i,k}} k_{j,k},
\]
then the partial derivative with respect to a learned grade \( q_{i,k} \) is:
\[
\frac{\partial \mathrm{Score}_{ij}}{\partial q_{i,k}} = \left( \la^{q_{i,k}} + q_{i,k} \la^{q_{i,k}} \ln \la \right) k_{j,k}.
\]
This expression reveals the exponential sensitivity of gradients to \( q_{i,k} \), especially for large values of \( \la \) or \( q_{i,k} \), and motivates several stabilization techniques.

To train the Exponentially Graded Transformer effectively, we propose the following strategies:

\begin{enumerate}[label=\roman*), ref=\roman*]
  \item \label{item:clipping} \textbf{Gradient Clipping}: Prevent gradient explosion by capping the gradient norm to a threshold \( \tau > 0 \). For a gradient vector \( \mathbf{g} \), we apply:
  \[
  \mathbf{g}' = \min\left(1, \frac{\tau}{\|\mathbf{g}\|_2}\right) \mathbf{g},
  \]
  when \( \|\mathbf{g}\|_2 > \tau \), leaving it unchanged otherwise.

  \item \label{item:warm_start} \textbf{Warm-Starting}: Initialize \( \w \) and \( \w_i \) using domain-informed priors (e.g., for polynomial degrees, use \( q_k = c \cdot k \), with \( c > 0 \)), to bias the model toward known hierarchical structures.

  \item \label{item:annealing} \textbf{Grade Annealing}: To avoid instability at early training stages, gradually increase the grading scale factor \( \la \) using a schedule:
  \[
  \la_t = 1 + (\la_{\max} - 1) \cdot \frac{t}{T},
  \]
  where \( t \) is the current training step and \( T \) is the total number of training steps. This \(\la_t\) is then substituted into all exponential maps \( \la^{q_k} \), ensuring consistent scaling.

  \item \label{item:coordination} \textbf{Multi-Head Grade Coordination}: To prevent diverging priorities among heads, we introduce a coordination regularizer:
  \[
  \cL_{\text{coord}} = \sum_{i=1}^h \left\| \w_i - \frac{1}{h} \sum_{j=1}^h \w_j \right\|_2^2,
  \]
  which is added to the total loss:
  \[
  \cL_{\mathrm{total}} = \cL + \gamma \| \w \|_2^2 + \gamma' \sum_{i=1}^h \| \w_i \|_2^2 + \gamma_{\text{coord}} \cL_{\text{coord}},
  \]
  with coordination regularization weight \( \gamma_{\text{coord}} > 0 \).
\end{enumerate}

These regularization and optimization techniques are crucial for maintaining stability during training, especially given the exponential nature of graded scaling. Without such controls, gradients involving \( \la^{q_k} \) can easily become unstable, either vanishing or exploding, particularly in deep or multi-headed architectures.

\subsection{Convergence and Gradient Stability}

Convergence and gradient stability are critical for ensuring the Exponentially Graded Transformer’s training process is robust, particularly given the exponential sensitivity introduced by learnable grading transformations.

\begin{thm}[Convergence]\label{thm:convergence}
Fix a grading vector \( \w \). If the loss function \( \ell \) is Lipschitz continuous, then gradient descent with sufficiently small step size converges to a stationary point of the Exponentially Graded Transformer’s loss.
\end{thm}

\begin{proof}
The Exponentially Graded Transformer \( \mathcal{EGT}_{\w, \la} \) is Lipschitz continuous by \cref{prop:noise_robustness}, with constant 
\[
 L \leq L_{\cT} (\la^{\w_{\max}})^{L h + 3},
 \]
  where \( q_{\max} = \max_{k=1,\ldots,d} q_k \). The exponentially graded loss \( \cL \) is Lipschitz continuous with constant \( L_\ell \cdot \max_k \la^{q_k} \), as \( \ell \) is Lipschitz with constant \( L_\ell \). The composite function \( \cL \circ \EGT \) has a Lipschitz gradient with constant proportional to \( L \cdot \max_k \la^{q_k} \). By classical gradient descent results \cite{nesterov2018lectures}, a step size \( \eta < 2/(L \cdot \max_k \la^{q_k}) \) ensures convergence to a stationary point.
\end{proof}

\begin{prop}[Gradient Stability]\label{prop:grad-stability}
Assume that \( \| \w_i \|_2, \| \bk_j \|_2 \leq C \) and that \( \left| \frac{\partial \ell}{\partial \hat{y}_{i,j}} \right| \leq L_\ell \). Then the gradient 
$
 \frac{\partial \cL_{\mathrm{total}}}{\partial q_k} 
$
  is Lipschitz continuous in \( q_k \), with constant proportional to \( \la^{\w_{\max}} \ln \la \).
\end{prop}

\begin{proof}
The form of the derivative is given by
\begin{equation}
\begin{split}
\frac{\partial \cL_{\mathrm{total}}}{\partial q_k} &	 = \sum_{i=1}^m \sum_{j=1}^d \left( \la^{q_j} \frac{\partial \ell}{\partial \hat{y}_{i,j}} \cdot \frac{\partial \hat{y}_{i,j}}{\partial q_k} + \ell(\hat{y}_{i,j}, y_{i,j}) \ln \la \cdot \la^{q_k} \delta_{j,k} \right) \\
& + 2\gamma q_k.
\end{split} 
\end{equation}
To verify Lipschitz continuity, compute the second derivative:
\begin{align}
\frac{\partial^2 \cL_{\mathrm{total}}}{\partial q_k^2} &= \sum_{i=1}^m \left( \la^{q_k} (\ln \la)^2 \ell(\hat{y}_{i,k}, y_{i,k}) \right.   \\ 
	&	\left.    + 2 \la^{q_k} \ln \la \cdot \frac{\partial \ell}{\partial \hat{y}_{i,k}} \cdot \frac{\partial \hat{y}_{i,k}}{\partial q_k}   + \la^{q_k} \frac{\partial^2 \ell}{\partial \hat{y}_{i,k}^2} \left( \frac{\partial \hat{y}_{i,k}}{\partial q_k} \right)^2 \right. \notag \\
& \left.  + \la^{q_k} \frac{\partial \ell}{\partial \hat{y}_{i,k}} \cdot \frac{\partial^2 \hat{y}_{i,k}}{\partial q_k^2} \right) + 2\gamma. \label{eq:second_deriv}
\end{align}
Assuming \( \ell \) is Lipschitz with constant \( L_\ell \), \( \left| \frac{\partial \ell}{\partial \hat{y}_{i,k}} \right| \leq L_\ell \), and \( \hat{y}_{i,k} \) depends on \( q_k \) through the output layer (\cref{subsec:architecture}), which is Lipschitz with constant \( \la^{\w_{\max}} L_{\text{out}} \). The dominant term is \( \la^{q_k} (\ln \la)^2 \ell(\hat{y}_{i,k}, y_{i,k}) \), bounded by \( m \la^{\w_{\max}} (\ln \la)^2 \cdot \text{const} \), yielding a Lipschitz constant proportional to \( \la^{\w_{\max}} \ln \la \).
\end{proof}

\begin{prop}[Training Stability for Learned Grades]
Assuming \( \|\w_i\|_2, \|\bk_j\|_2 \leq C \), the training process with learned \( \w \) is stable if the learning rate for \( q_k \) satisfies \( \eta_q < 1/(\la^{\w_{\max}} \ln \la) \).
\end{prop}

\begin{proof}
From \cref{prop:grad-stability}, the gradient \( \frac{\partial \cL_{\mathrm{total}}}{\partial q_k} \) is Lipschitz with constant proportional to \( \la^{\w_{\max}} \ln \la \). For gradient descent updates 
\begin{equation}
 q_k \leftarrow q_k - \eta_q \frac{\partial \cL_{\mathrm{total}}}{\partial q_k},
 \end{equation}
  stability requires \( \eta_q < 1/L_g \), where \( L_g \propto \la^{\w_{\max}} \ln \la \). Thus, choosing \( \eta_q < 1/(\la^{\w_{\max}} \ln \la) \) ensures stable updates, preventing divergence due to exponential sensitivity in \( \la^{q_k} \).
\end{proof}

\begin{prop}[Sensitivity to \(\la\)]
The Exponentially Graded Transformer’s training is sensitive to the choice of \( \la \), with the loss gradient’s magnitude scaling as \( O(\la^{\w_{\max}} \ln \la) \). A bounded \( \la \leq \la_{\max} \) ensures stable optimization.
\end{prop}

\begin{proof}
The gradient \( \frac{\partial \cL_{\mathrm{total}}}{\partial q_k} \) includes terms like 
\begin{equation}
 \la^{q_k} \ln \la \cdot \ell(\hat{y}_{i,k}, y_{i,k}),
 \end{equation}
  scaling with \( \la^{\w_{\max}} \ln \la \). From \cref{prop:noise_robustness}, the Exponentially Graded Transformer’s Lipschitz constant scales as \( (\la^{\w_{\max}})^{L h + 3} \). Large \( \la \) amplifies gradients, risking instability. Bounding 
  \begin{equation}
   \la \leq \la_{\max} 
   \end{equation}
    (e.g., \( \la_{\max} = 2 \)) limits the gradient magnitude, ensuring stable optimization with \[ \eta_q < 1/(\la_{\max}^{\w_{\max}} \ln \la_{\max}).\]
\end{proof}

\subsection{EGT-Specific Training Challenges}
\label{subsec:gt_training_challenges}

The Exponentially Graded Transformer introduces novel training challenges due to its exponential scaling and multi-head grading structure. In contrast to earlier models \cite{2024-2, sh-89}, where grading tuples were fixed, here both the global grade vector \( \w = (q_0, \ldots, q_{d-1}) \) and the per-head vectors \( \w_i = (q_{i,0}, \ldots, q_{i,d_k-1}) \) are treated as learnable parameters. This improves adaptability but increases the risk of instability in optimization. The following algorithm summarizes a robust training procedure tailored to EGT.

\begin{algocf}[Exponentially Graded Training]
\label{alg:gt_training}

\begin{algorithmic}
  \STATE
  
  \STATE \textbf{Input}: Training data, initial model parameters \( \theta \), grade tuples \( \w \), \( \w_i \), initial \( \la = 1 \), target \( \la_{\max} \), total steps \( T \), gradient threshold \( \tau \), regularization weights \( \gamma, \gamma', \gamma_{\text{coord}} \).
  
  \STATE \textbf{Note}: All grades \( \w, \w_i \) are learnable.\\
  
  \STATE \textbf{Initialize}:
  
  \STATE \quad Set \( \w = (q_0, \ldots, q_{d-1}) \) using domain-informed values (e.g., \( q_k = c \cdot k \), \( c > 0 \)),
  \STATE \quad Set \( \w_i = (q_{i,0}, \ldots, q_{i,d_k-1}) \) similarly for each attention head \( i = 1,\ldots,h \). \\
  
  \FOR{each training step \( t = 1, \ldots, T \)}
    \STATE Update \( \la_t = 1 + (\la_{\max} - 1) \cdot \frac{t}{T} \).
    \STATE Compute loss components:
      \begin{align*}
        \cL_{\text{coord}} &= \sum_{i=1}^h \left\| \w_i - \frac{1}{h} \sum_{j=1}^h \w_j \right\|_2^2, \\
        \cL_{\text{total}} &= \cL + \gamma \| \w \|_2^2 + \gamma' \sum_{i=1}^h \| \w_i \|_2^2 + \gamma_{\text{coord}} \cL_{\text{coord}}.
      \end{align*}
    \STATE Compute gradients \( \nabla \cL_{\text{total}} \) and apply clipping:
      \[
      \mathbf{g}' = \begin{cases}
      \mathbf{g}, & \text{if } \|\mathbf{g}\|_2 \leq \tau, \\
      \frac{\tau}{\|\mathbf{g}\|_2} \cdot \mathbf{g}, & \text{otherwise}.
      \end{cases}
      \]
    \STATE Estimate \[ \w_{\max} = \max \left( \max_k q_k, \max_{i,k} q_{i,k} \right) \]
    
    \STATE Use learning rate \( \eta_q < 1 / (\la_t^{\w_{\max}} \ln \la_t) \) for grade updates; update \( \theta, \w, \w_i \) using Adam.
  \ENDFOR
  \STATE \textbf{Output}: Optimized parameters \( \theta, \w, \w_i \).
\end{algorithmic}
\end{algocf}

The components of Algorithm~\ref{alg:gt_training} are designed to address the following key training challenges:

\begin{enumerate}[label=\roman*), ref=\roman*]
  \item \textbf{Grade Initialization}: Domain-informed initialization (e.g., \( q_k = c \cdot k \)) improves convergence and prevents poor early gradients due to randomly scaled inputs.

  \item \textbf{Tuning \(\la\)}: Large values of \( \la \) amplify important features (\cref{prop:noise_robustness}) but risk instability via terms like \( \la^{q_k} \). Annealing \( \la_t \) from 1 to \( \la_{\max} \) mitigates this risk while preserving expressivity.

  \item \textbf{Multi-Head Coordination}: Independent grade vectors \( \w_i \) enhance expressivity (\cref{prop:head_diversity}) but can diverge. The regularization term \( \cL_{\text{coord}} \) penalizes deviations from the headwise mean, promoting coherence among heads.

  \item \textbf{Gradient Control}: Exponentially graded mappings produce gradients sensitive to \( q_k \), requiring gradient clipping to prevent numerical overflow. The learning rate for grades is adaptively bounded by \( \la_t^{\w_{\max}} \ln \la_t \) to ensure stability.
\end{enumerate}

Together, these procedures ensure stable and effective training of EGT models. Empirical validation on tasks such as symbolic regression and structured prediction (cf.~\cref{sec-8}) will further illustrate the benefits of adaptive grading over fixed architectures \cite{2024-2, sh-89}.
 
\section{Potential Applications}
\label{sec-8}

The graded transformer embeds hierarchical priors via LGT and EGT, enabling principled learning across structured domains. This section outlines potential impact in algebraic geometry, physics, natural language processing (NLP), biological sequence analysis, cross-domain settings, and emerging areas such as graph learning and financial modeling. Each domain exhibits natural grading—via degree, scale, structure, or functional relevance—which the graded transformer exploits to enhance attention efficiency (\cref{prop:attention_rank}), reduce sample complexity (\cref{prop:lgt_vc_dimension}), and improve interpretability. Linear grading suits moderate hierarchies, while exponential grading amplifies deep structures. Mathematical examples illustrate how grading is applied, and new applications highlight the graded transformer's versatility, while acknowledging challenges in grade design.

\subsection{Algebraic Geometry}
\label{subsec:alg_geom}

Graded structures are central to algebraic geometry, from polynomial rings to moduli spaces and weighted projective varieties. The graded transformer aligns with these by emphasizing basis elements according to algebraic degree or weighted monomial structure, with learnable grades adapting to complex geometries.
\begin{enumerate}[label=\roman*), ref=\roman*]
  \item \label{item:alg_geom_moduli} Modeling moduli spaces and isogenies, where graded invariants govern geometry~\cite{sh-93, sh-91, 2024-3}; exponential grading amplifies high-degree terms.
  \item \label{item:alg_geom_zeta} Computing zeta functions and point counting on hypersurfaces, scaling monomials via grading to reduce complexity (\cref{prop:lgt_vc_dimension}).
\end{enumerate}

Mapping discrete degrees to continuous grades remains nontrivial; this is mitigated by learning \(\w\) (\cref{subsec:gt_training_challenges}).

\subsection{Physics}
\label{subsec:physics}
Many physical systems are multiscale, from microscopic energy levels to macroscopic dynamics. Grading emphasizes dominant features consistent with physical intuition (quantum mechanics, fluid dynamics, cosmology).
\begin{enumerate}[label=\roman*), ref=\roman*]
  \item \label{item:physics_quantum} Spectral prediction in quantum systems, scaling by energy eigenvalues to emphasize high-energy orbitals~\cite{sh-89}; linear grading for moderate scales.
  \item \label{item:physics_turbulence} Turbulence and phase transitions, emphasizing low-wavenumber structures via inverse-frequency grading for stability (\cref{prop:noise_robustness}).
\end{enumerate}

Continuous spectra can leverage learned grades; see \cref{subsec:gt_training_challenges}.

\subsection{Natural Language Processing}
\label{subsec:nlp}
Language exhibits hierarchical structure; words and syntactic roles have varying semantic weight. The graded transformer adjusts attention via learnable grades, encoding structural and sequential hierarchies (\cref{prop:positional_bias}).
\begin{enumerate}[label=\roman*), ref=\roman*]
  \item \label{item:nlp_parsing} Syntactic parsing and semantic role labeling, scaling heads via grading for efficiency (\cref{prop:attention_rank}).
  \item \label{item:nlp_dialogue} Dialogue understanding and question answering, elevating intent-bearing tokens with reduced VC dimension (\cref{prop:lgt_vc_dimension}).
\end{enumerate}

Adapting grades to syntactic conventions benefits from structured corpora and learned \(\w_i\).

\subsection{Biological Sequence Analysis}
\label{subsec:bio_seq}

Biological sequences contain functionally significant elements embedded in noise. Grading scales these elements, improving sample efficiency (\cref{prop:lgt_vc_dimension}) in genomics and proteomics.
\begin{enumerate}[label=\roman*), ref=\roman*]
  \item \label{item:bio_gene} Gene prediction and variant effect modeling, emphasizing coding regions via exponential grading for deep hierarchies.
  \item \label{item:bio_protein} Protein modeling and metagenomics, weighting active sites with robustness guarantees (\cref{prop:noise_robustness}).
\end{enumerate}

\subsection{Cross-Domain Applications}
\label{subsec:cross_domain}

Grading abstracts importance across domains, enabling transfer learning on structured data.

\begin{enumerate}[label=\roman*), ref=\roman*]
  \item \label{item:cross_symbolic} Pretraining on symbolic tasks (e.g., Gröbner-basis learning) followed by fine-tuning on NLP or genomics.
  \item \label{item:cross_data_fabrics} Encoding data fabrics as 4D graded structures for flow-aware learning.
\end{enumerate}

\subsection{Robotics}
Robotics systems must balance safety and efficiency in dynamic environments. Graded architectures provide a principled mechanism for prioritizing safety-critical features over throughput or energy goals. For example, collision-avoidance signals from proximity sensors receive higher grades than energy-optimization objectives, ensuring obstacle detection dominates decisions. 

In human--robot interaction, high-grade features tied to human presence or gestures override lower-grade efficiency goals, aligning behavior with safety and ethical imperatives. Recent advances, such as logically constrained transformers for safe robotic planning~\cite{generalrobotics2024}, demonstrate how grading can integrate with temporal logic to enforce safety constraints in real-time planning. Furthermore, diffusion-based models like the Robotics Diffusion Transformer~\cite{acm2025} could benefit from exponential grading to amplify hierarchical action sequences in foundation models for robotics.

\subsection{Automotive}

In ADAS and autonomous vehicles, certain decisions must carry mathematically provable priority. For example, ``do not hit a pedestrian'' should outrank ``maintain optimal speed.'' Graded transformers encode such hierarchies by amplifying safety-critical features in attention and decision layers, making them dominant in the output. 

This graded prioritization supports interpretable and verifiable AI, offering algebraic guarantees that safety constraints are not overridden by secondary objectives (efficiency, comfort), and thereby aiding regulatory certification. Transformer-based object detection models for self-driving cars~\cite{ieee2024} can be enhanced with linear grading to prioritize dynamic obstacles, while generative frameworks for pre-crash trajectories~\cite{sciencedirect2025} leverage exponential grading for multi-scale temporal dependencies in complex driving scenarios.

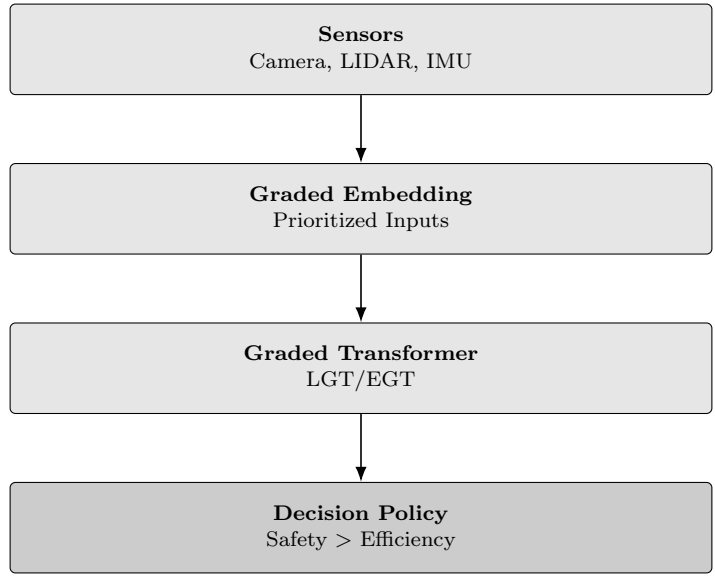
\begin{figure}[t]
\centering

\usetikzlibrary{positioning,arrows.meta}
\tikzset{
  gt/block/.style={
    draw,
    rounded corners=2pt,
    minimum width=0.58\columnwidth, 
    minimum height=1.2cm,
    inner sep=6pt,
    fill=black!10,
    align=center,
    font=\footnotesize
  },
  gt/arrow/.style={
    -{Latex[length=2mm]},
    line width=0.6pt
  }
}

\begin{tikzpicture}[node distance=0.9cm]
  \node[gt/block] (sensors) {\textbf{Sensors}\\Camera, LIDAR, IMU};
  \node[gt/block, below=of sensors] (embed) {\textbf{Graded Embedding}\\Prioritized Inputs};
  \node[gt/block, below=of embed] (transformer) {\textbf{Graded Transformer}\\LGT/EGT};
  \node[gt/block, fill=black!20, below=of transformer] (policy) {\textbf{Decision Policy}\\Safety $>$ Efficiency};

  \draw[gt/arrow] (sensors) -- (embed);
  \draw[gt/arrow] (embed) -- (transformer);
  \draw[gt/arrow] (transformer) -- (policy);
\end{tikzpicture}

\caption{Graded transformer pipeline for safety-critical systems. High-grade safety features dominate lower-priority objectives throughout the stack.}
\label{fig:graded_pipeline_vertical}
\end{figure}

\subsection{Integration with Symbolic Rules}

A key advantage of graded transformers is their neuro-symbolic nature, allowing seamless integration of continuous, sensor-driven learning with discrete symbolic rules. In automotive contexts, symbolic constraints (traffic laws, right-of-way, ethical directives) can be assigned explicit grades to ensure compliance despite perceptual uncertainty. 

In robotics, symbolic goals such as ``do not harm humans'' are embedded as high-grade features, guaranteeing dominance in learning and inference. This hybrid framework bridges numerical optimization with logical reasoning, yielding a transparent and mathematically rigorous approach to safety-critical decision making. Neuro-symbolic frameworks, such as those integrating reasoning with deep learning for enhanced decision-making in autonomous robotics~\cite{researchgate2024}, highlight how graded priors can enforce safety in agile control systems~\cite{yue2024}, while applications in traffic domains emphasize predictive tasks like safety and perception~\cite{neurosymbolic2025}.

\subsection{Additional Applications}
\label{subsec:neq_applications}

The graded approach extends to further hierarchical domains.

\begin{enumerate}[label=\roman*), ref=\roman*]
  \item \label{item:neq_graph} Graph neural networks, grading nodes by centrality for community detection or molecular modeling.
  \item \label{item:neq_finance} Financial time-series forecasting, grading significant market events for volatility prediction.
\end{enumerate}

These examples underscore versatility: learned grades adapt to task hierarchies and enhance efficiency and interpretability. Open challenges remain in grade design and transferability; broad empirical validation across these tasks is an important direction for future work.

\begin{rem}
Certain aspects of the graded transformer framework are the subject of recent patent applications by the author, highlighting its potential for practical deployment in domains such as NLP, robotics, automotive industry, and financial cryptography.
\end{rem}

\begin{rem}
Several components of the graded transformer framework are the subject of recent patent applications, underscoring its potential for real-world deployment in NLP, robotics, automotive, and financial cryptography.
\end{rem}


\bibliography{sh-95.bib}

\end{document}